\documentclass[review]{elsarticle}
\usepackage{lineno,hyperref}
\def\etal{~\emph{et al}.~}
\usepackage{multirow}
\usepackage{color}
\usepackage{multirow}
\usepackage{booktabs}
\usepackage{bbm}
\usepackage[ruled,vlined]{algorithm2e}
\usepackage{array}
\usepackage{amsmath}
\usepackage{amssymb, bm} 
\usepackage{graphicx}

\usepackage{geometry}
\modulolinenumbers[5]

\begin{document}

\begin{frontmatter}

\title{Sparsely-Labeled Source Assisted Domain Adaptation}


\author[dut]{Wei~Wang}
\author[dut]{Zhihui~Wang\corref{cor}}
\cortext[cor]{
	Corresponding author:
}
\ead{zhwang@dlut.edu.cn}
\author[dut]{Yuankai~Xiang}
\author[dut]{Jing~Sun}
\author[dut]{Haojie~Li}
\author[dmz]{Fuming~Sun}
\author[iup]{Zhengming~Ding}

\address[dut]{DUT-RU International School of Information Science $\&$ Engineering\\ Dalian University of Technology, Dalian 116000, P.R. China}
\address[dmz]{School of Information $\&$ Communication\\ Dalian Minzu University, Dalian 116600, P.R. China}
\address[iup]{Department of Computer, Information and Technology, Purdue School of Engineering and Technology\\ Indiana University-Purdue University Indianapolis, Indianapolis IN 46202, USA}

\begin{abstract}
Domain Adaptation (DA) aims to generalize the classifier learned from the source domain to the target domain. Existing DA methods usually assume that rich labels could be available in the source domain. However, there are usually a large number of unlabeled data but only a few labeled data in the source domain, and how to transfer knowledge from this sparsely-labeled source domain to the target domain is still a challenge, which greatly limits their application in the wild. This paper proposes a novel Sparsely-Labeled Source Assisted Domain Adaptation (SLSA-DA) algorithm to address the challenge with limited labeled source domain samples. Specifically, due to the label scarcity problem, the projected clustering is first conducted on both the source and target domains, so that the discriminative structures of data could be leveraged elegantly. Then the label propagation is adopted to propagate the labels from those limited labeled source samples to the whole unlabeled data progressively, so that the cluster labels are revealed correctly. Finally, we jointly align the marginal and conditional distributions to mitigate the cross-domain mismatch problem, and optimize those three procedures iteratively. However, it is nontrivial to incorporate those three procedures into a unified optimization framework seamlessly since some variables to be optimized are implicitly involved in their formulas, thus they could not promote to each other. Remarkably, we prove that the projected clustering and conditional distribution alignment could be reformulated as different expressions, thus the implicit variables are revealed in different optimization steps. As such, the variables related to those three quantities could be optimized in a unified optimization framework and facilitate to each other, to improve the recognition performance obviously. Extensive experiments have verified that our approach could deal with the challenge in SLSA-DA setting, and best performances could be achieved on different real-world cross-domain visual recognition tasks.
\end{abstract}


\end{frontmatter}


\section{Introduction}

\begin{figure}[h]
	\begin{center}
		\includegraphics[width=0.8\linewidth,height=0.3\textheight]{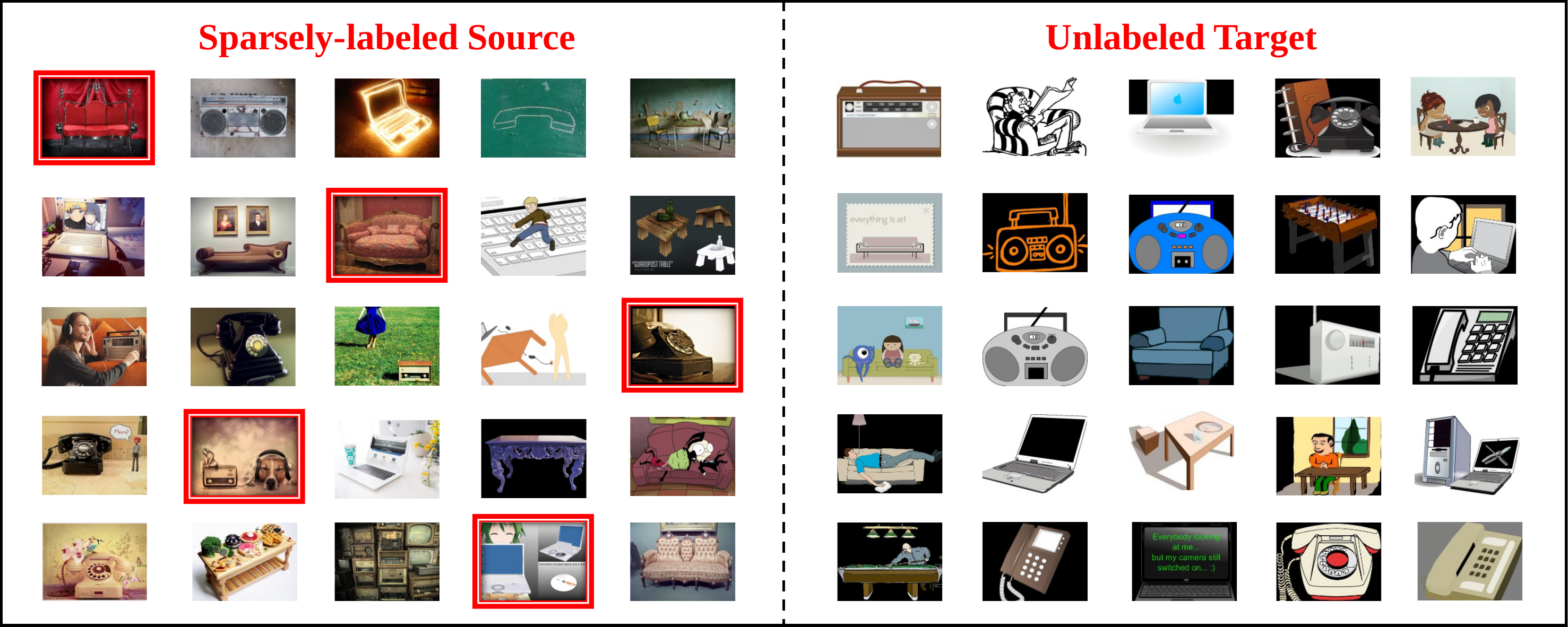}
	\end{center}
	\vspace{-10pt}
	\caption{The sparsely-labeled source assisted domain adaptation problem, where the samples in red bounding boxes are labeled source, and the others are unlabeled.}
	\label{fig1}
\end{figure}

\par Domain Adaptation (DA) has received much attention in recent years as it offers possibility to generalize the classifier trained on one domain to another domain, where the observed data sampled from those two domains are usually coming from different distributions~\cite{STL1}. For example, in visual recognition, the data instances of those two domains are usually originated from different environments, sensor types, resolutions, and view angles, so that they would follow very discrepant distributions~\cite{PR6}. It is impractical to annotate sufficient data for each domain since labeling data is labor intensive and expensive. Therefore, it is necessary to apply the DA techniques to exploit invariant features across different domains, so that well-labeled source knowledge could be transferred to target domain, then the labeling consumption is mitigated. Recently, DA has made remarkable progress in cross-domain hyperspectral image classification~\cite{PR2}, human action recognition~\cite{PR5}, etc.

\par However, the performances of traditional DA usually significantly rely on label quality or richness in the source domain, which is restricted in applications in the wild as we still have to seek a better-labeled and higher-quality source domain~\cite{WSDA, WOSDA}. Label quality and its sufficiency are both important in context of domain adaptation (DA), especially for deep learning DA frameworks~\cite{PR3}. In some real applications, it may not be easy for users to label data samples as correctly and sufficiently as possible, since they often struggle with a very complicated and large dataset. For example, there exist some data points, which are ambiguous between different categories, or require high level professional technologies. Therefore, those data samples are easily partial wrongly-labeled when they commit to annotate the whole dataset, or highly sparsely-labeled when they only label a handful of samples to reduce the labeling consumption as much as possible. Moreover, due to the dataset is large, it is more challenging to guarantee the label quality or its richness, especially the deep learning DA frameworks which often requires vast amounts of source domain data. Therefore, the provided poor labeled dataset, to be regarded as source domain, have a great impact on training processes of DA models, since incorrect or unknown knowledge of source domain will cause unexpectantly and heavily negative transfer~\cite{STL1}. Therefore, it is essential to study the situation when the given source domain is poor-labeled, either partial wrongly-labeled (label quality) or highly sparsely-labeled (label richness). 

\par In practical applications, it may not be possible to access a significant amount of labeled data, especially with the dramatic increase of data in deep learning models. Therefore, it is essential to boost positive transfer for a newly unlabeled target domain using a poor-labeled source domain. Many examples in knowledge engineering could be found where this situation can truly be beneficial. One example is the problem of sentiment classification, where our task is to automatically classify the reviews on a product or scores on a visual image. For this classification task, we need to first collect many products or visual images and annotate them using the given reviews or scores. However, labeling them is very labor-intensive and mind-numbing for users, since some data samples are too ambiguous and have no significant divergences between various categories, especially the scores on visual images. Therefore, the label quality or richness is poor when we commit to label them all or only annotate a handful of them to reduce labeling consumption. 

\par After that, we would utilize this poor dataset, to be regarded as the source domain, to train a classifier. Since the distribution of data among different types of products or visual images can be very different, to maintain good classification performance, we need to recollect the source domains in order to train the review/score-classification models for each kind of products/visual images. However, this data-labeling process can be also very expensive to do. To further reduce the effort for annotating reviews/scores for various products/visual images, we may want to adapt a classification model that is trained on some products/visual images (poor-labeled), which could be directly applied to make prediction for other types of products/visual images. As another example, we can consider the data information of users collected from different supermarkets (e.g., Walmart and Amazon, etc.), which is updated every day and often in large quantities, thus it may be also impossible to guarantee the label quality and its richness. Therefore, it is very challenging for us to utilize the poor-labeled data information of users collected from a supermarket (e.g., Walmart) to exploit the interest of users from another different supermarket (e.g., Amazon). In such cases, the proposed sparsely-labeled source assisted domain adaptation can save a significant amount of labeling effort. 

\par To this end, Weakly-Supervised Domain Adaptation (WSDA) is proposed to address the challenge that the source domain contains noises in labels, features, or both~\cite{WSDA}. However, they only focus on the label quality problem, and do not further explore that the source labels are insufficient severely. A more realistic setting, Sparsely-Labeled Source Assisted Domain Adaptation (SLSA-DA), is therefore proposed in this paper, to further mitigate the labeling consumption, where only a sparsely-labeled source domain is available without any target labels. Notably, this paper assumes that the target domain is completely unlabeled to increase the difficulty of our work since previous DA work indicates that the unsupervised DA~\cite{PR1, PR8} is more challenging than the semi-supervised one~\cite{PR4, PR7}. Moreover, we aim to enable the proposed model more general since it is a special and more simple case when there exists at least one example of the class in the target domain. For example, on the Office-Home dataset, $60\%$ source labels are available correctly in WSDA setting, while only $7.2\%\sim15.6\%$ in SLSA-DA scenario. As shown in Fig.~\ref{fig1}, there are numerous unlabeled data but a few labeled ones, then we need to utilize this sparsely-labeled source domain to assist recognition for target domain.

\par In order to address the challenge of SLSA-DA, our aim is not only to fight off the label insufficiency issues in the source domain, but also to mitigate the domain shift across the source and target domains. It is essential for DA to study this new SLSA-DA scenario, which could implement knowledge transfer with lowest labeling cost than most existing approaches. Specifically, SLSA-DA introduces two challenges. (1) It is still significant to alleviate the influence of distributional shift across different domains as presented in previous DA methods. (2) Moreover, it is nontrivial to train a well-structured classifier since only limited source labels are available.

\par
\begin{figure}[h]
	\begin{center}
		\includegraphics[width=0.8\linewidth,height=0.35\textheight]{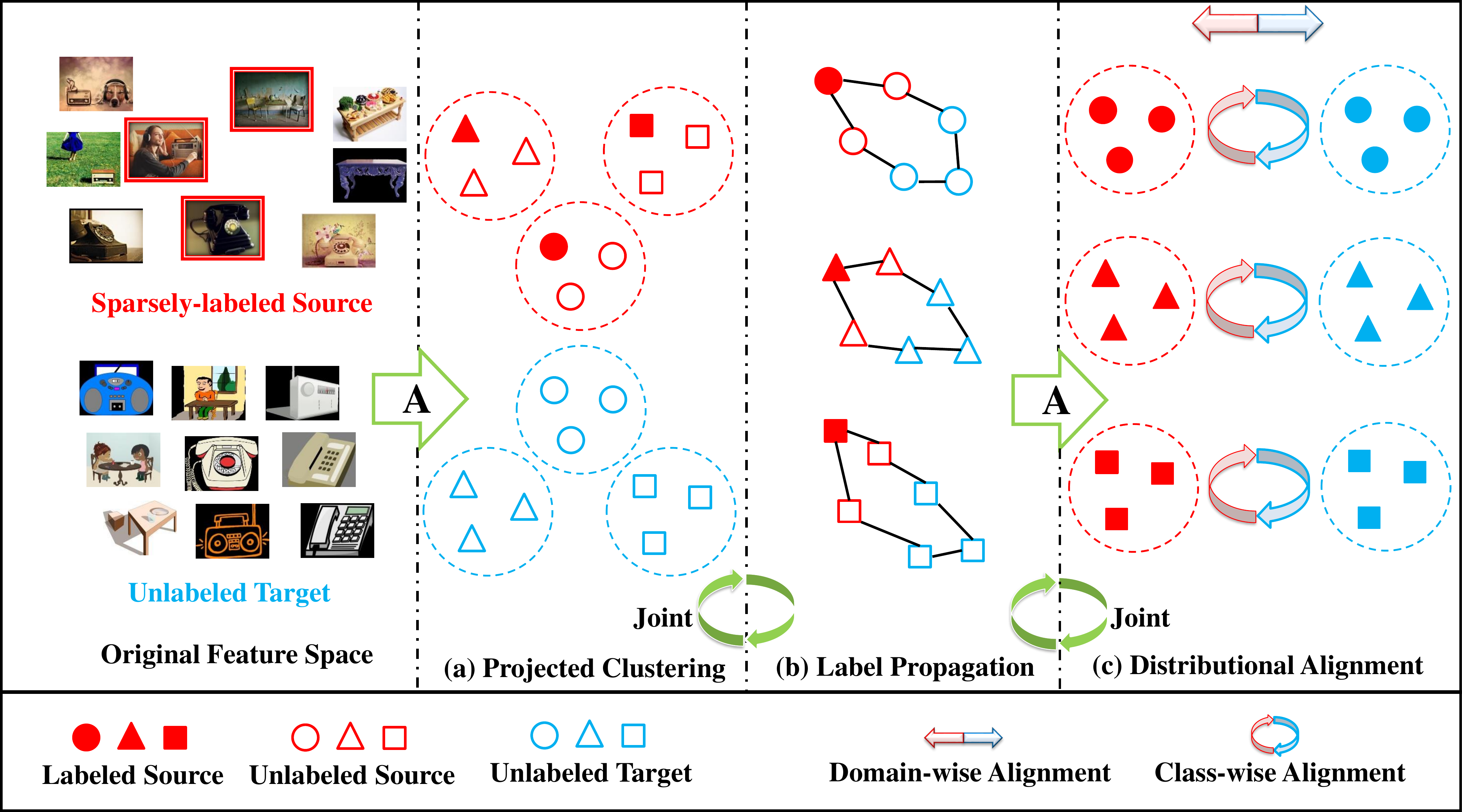}
	\end{center}
	\vspace{-10pt}
	\caption{The overview of the proposed approach.}
	\label{fig2}
\end{figure}

 Due to the label scarcity problem in SLSA-DA setting, we carry the semi-supervised projected clustering on the source domain using a few labeled source instances, while unsupervised on the target domain, so that the discriminative structures of data could be discovered desirably, i.e., data samples from the same cluster are assembled tightly (i.e., Fig.~\ref{fig2}~(a)). Although the cluster labels of source domain can be consistent with ground-truth labels, it is uncertain in the target domain since no supervised information is provided. Therefore, the label propagation method \cite{LP2} is adopted to propagate the source limited labels to the source and target unlabeled instances simultaneously, so that the target cluster labels are revealed as correctly as possible (i.e., Fig.~\ref{fig2}~(b)). Once their labels are uncovered, we can jointly align the marginal and conditional distributions across different domains using the methods of Maximum Mean Discrepancy (MMD)~\cite{MMD} and class-wise MMD~\cite{JDA} (i.e., Fig.~\ref{fig2}~(c)). In order to refine the final recognition performance progressively and enable different steps facilitate to each other, we iteratively conduct those three procedures in a few times. 

\par However, it is nontrivial to integrate the projected clustering, label propagation and distributional alignment as a unified optimization framework, since some variables to be optimized are implicitly involved in their formulas, thus they could not promote to each other. To be specific, the construction of class-wise MMD implicitly contains the variables related to cluster centroids, but those variables in the projected clustering should be implicit when we optimize the projection matrix. Existing DA models are usually formulated with the label prediction and distributional alignment, and separate them as different steps \cite{GAKT}. Therefore, they will fail to take advantage of each other's merits and promote to each other. In contrast, this paper further considers the projected clustering so that the model is robust to the label scarcity problem as we respect the discriminative structures of data. Moreover, we prove that the class-wise MMD could be rewritten as the cluster-wise MMD when we optimize the variables related to cluster centroids, while the projected clustering could be reformulated as the intra-class scatter minimization \cite{UTR} when we optimize the shared projection matrix. Therefore, we could couple those three quantities together and benefit them to each other in an effective optimization manner.

\par The main contributions of our work are two-folds:

\begin{itemize}
	\item We first introduce a new DA scenario, called Sparsely-Labeled Source Assisted Domain Adaptation, which is more realistic as it requires a few labeled source data while is under insufficient exploration so far.
	\item  We propose a unified framework to jointly seek cluster centroids, source and target labels, and domain-invariant features. Then, we construct an optimization strategy to solve the objective function efficiently.
\end{itemize}

The rest of the paper is organized as follows. The related works are reviewed in Section 2. In Section 3, we propose the model and SLSA-DA algorithm. The experimental evaluations are discussed in Section 4. Finally, we conclude this paper in Section 5.

\section{Related Work}

\par Traditional DA aims to employ previous labeled source domain data to boost the task in the target domain. However, they usually assume the source and target domains share an identical label space, known as Closed Set Domain Adaptation (CSDA). Recently, an increasing number of new domain adaptation scenarios have been proposed to compensate for different challenges in practical application, such as Partial Domain Adaptation (PDA), Open Set Domain Adaptation (OSDA), and Universal Domain Adaptation (UDA). PDA transferred a learner from a big source domain to a small target domain, and the label set of the source domain is supposed to be large enough to contain the target label set~\cite{PDA1}. By contrast, OSDA was proposed to deal with the challenge that the target domain contains unknown classes, which are not observed in the source domain~\cite{OSDA1}. Furthermore, for a given source label set and a target label set, UDA required no prior knowledge on the label sets, where they may contain a common label set and hold a private label set respectively~\cite{UDA}.

\par All of aforementioned works have shown great improvements in the performance of knowledge transfer due to the available substantial amount of high-quality labeled data in the source domain. Therefore, recent research has set about following with interest the weakly supervised DA scenario. For instance, Tan\etal~\cite{WOSDA} proposed a Collaborative Distribution Alignment (CDA) method for a Weakly Supervised Open-Set Domain Adaptation (WSOSDA), where both domains are partially labeled and not all classes are shared between these two domains. In contrast, Long\etal~\cite{WSDA} proposed a Transferable Curriculum Learning (TCL) approach to address the challenge of sample noises of the source domain in a Weakly Supervised Close-Set Domain Adaptation (WSCSDA). However, their settings still require enough labeled instances either in the source or target domains. In order to further mitigate the intensive labeling expenses, we propose a more realistic DA paradigm, called Sparsely-Labeled Source Assisted Domain Adaptation, which requires only a few source labels and a satisfactory performance could be warranted through a proposed unified framework. In order to highlight the contributions in this paper and make the model simpler, SLSA-DA assumes the source and target label sets are the same, and the source labeled instances are sparsely located in each class. To the best of our knowledge, our work is the first attempt to deal with this sparsely-labeled WSCSDA scenario.

\par Recent DA methods follow a mainstream approach which is based on the feature adaptation (FDA). FDA aims to extract a shared subspace, where the distributions of the source and target data are drawn close by explicitly minimizing some predefined distance metrics, e.g., Bregman Divergence~\cite{BD}, Geodestic Distance~\cite{GFK}, Wasserstein Distance~\cite{WD} and Maximum Mean Discrepancy (MMD)~\cite{MMD}. The most popular distance is MMD due to its simplicity and solid theoretical foundations~\cite{TRSC}. Pan \etal~\cite{TCA} proposed the Transfer Component Analysis (TCA) to align the marginal distribution across domains using MMD. Long \etal~\cite{JDA} proposed class-wise MMD to further reduce the conditional distribution difference between the two domains. Furthermore, SCA~\cite{SCA}, JGSA~\cite{JGSA}, VDA~\cite{VDA} constructed the class scatter matrix of source domain to preserve its discriminative information. This paper also utilizes the MMD and class-wise MMD to jointly align the marginal and conditional distributions across the source and target domains. Moreover, we prove that the projected clustering process is equivalent to boost the intra-class compactness when the projection is optimized. Therefore, the learned features from the proposed model are domain-invariant and discriminative, simultaneously.

\par It is noteworthy that the methods mentioned above require a strong assumption that rich labels are available in the source domain. Moreover, they optimize the target labels in a separate step along with the domain-invariant feature learning, thus they may fail to benefit to each other in an effective manner~\cite{GAKT}. Different from them, this paper incorporates the projected clustering, label propagation and distributional alignment into a unified optimization framework seamlessly, and jointly optimize cluster centroids, source and target labels and domain-invariant features, where only a few source labels are available.

\section{Methodology}

\par In this section, we present our proposed model and its optimization strategy in detail.
\subsection{Problem Definition}

\par We begin with the definitions of terminologies. $\textbf{\textit{X}}_s\in{\mathbf{R}^{m\times{n_s}}}$ (resp. $\textbf{\textit{X}}_t\in{\mathbf{R}^{m\times{n_t}}}$) denotes the source (resp. target) domain data, where $n_s$ (resp. $n_t$) is the number of samples and $\textit{m}$ is the dimension of data instance. In the proposed SLSA-DA setting, there are a few source labels while no target labels at all, i.e., $D_s=D_s^l\cup{D_s^u}=\{(\textbf{\textit{x}}_i,\textbf{\textit{y}}_i)\}_{i=1}^{n_{sl}}\cup\{(\textbf{\textit{x}}_j)\}_{j=1}^{n_{su}},n_{sl}+n_{su}=n_s,n_{sl}\ll{n_s}$, $D_t=\{(\textbf{\textit{x}}_k)\}_{k=1}^{n_t}$, where $\textbf{\textit{x}}_i,\textbf{\textit{x}}_j,\textbf{\textit{x}}_k\in{\mathbf{R}^m}$ and the one-hot label $\textbf{\textit{y}}_i\in{\mathbf{R}^C}$ ($C$ is the number of classes). 

\par Moreover, we assume that the source and target domains follow the same feature space and label space, while the marginal and conditional distributions are different due to the dataset shift. Our aim is to find a projection $\textbf{\textit{A}}\in{\mathbf{R}^{m\times{k}}}$ to map $D_s$ and $D_t$ into a shared subspace, where those two distributional differences could be explicitly reduced. Then their new representations are $\textbf{\textit{Z}}_s=\textbf{\textit{A}}^{\top}\textbf{\textit{X}}_s,\textbf{\textit{Z}}_t=\textbf{\textit{A}}^{\top}\textbf{\textit{X}}_t,\textbf{\textit{Z}}_s\in{\mathbf{R}^{k\times{n_s}}},\textbf{\textit{Z}}_t\in{\mathbf{R}^{k\times{n_t}}}$.

\subsection{Projected Clustering}

\par The projected clustering aims to jointly optimize the cluster centroids and cluster labels in an embedding space, so that the data instances from the same clusters could be grouped together~\cite{UTR}. Since only limited source labels are available in SLSA-DA scenario, we propose to utilize a semi-supervised projected clustering in the source domain, while unsupervised setting in the target domain. Therefore, the discriminative structures of data could be exploited with these limited source labels. The loss of projected clustering $\textbf{\textit{L}}_{pc}$ is defined as follows:  

\begin{equation}
\begin{array}{lr}
\textbf{\textit{L}}_{pc}=||\textbf{\textit{A}}^{\top}\textbf{\textit{X}}_s-\textbf{\textit{A}}^{\top}\textbf{\textit{X}}_s\textbf{\textit{G}}_s\textbf{\textit{F}}_s^{\top}||_F^2+||\textbf{\textit{A}}^{\top}\textbf{\textit{X}}_t-\textbf{\textit{A}}^{\top}\textbf{\textit{X}}_t\textbf{\textit{G}}_t\textbf{\textit{F}}_t^{\top}||_F^2
\quad s.t. \quad \textbf{\textit{G}}_s\succeq{\textbf{\textit{0}}},\textbf{\textit{G}}_t\succeq{\textbf{\textit{0}}},\textbf{\textit{F}}_s(:,1:n_{sl})=\textbf{\textit{Y}}_s^l,
\end{array}
\label{eq1}
\end{equation}

\noindent where $\textbf{\textit{F}}_s\in{\mathbf{R}^{n_s\times{C}}}$, $\textbf{\textit{F}}_t\in{\mathbf{R}^{n_t\times{C}}}$ are the one-hot cluster labels for the source and target domains, respectively. According to~\cite{UTR}, the source and target cluster centroids could be computed as $\textbf{\textit{A}}^{\top}\textbf{\textit{X}}_s\textbf{\textit{G}}_s$, $\textbf{\textit{A}}^{\top}\textbf{\textit{X}}_t\textbf{\textit{G}}_t$, where $\textbf{\textit{G}}_{(s/t)}=\textbf{\textit{F}}_{(s/t)}(\textbf{\textit{F}}_{(s/t)}^{\top}\textbf{\textit{F}}_{(s/t)})^{-1}$. Eq.(\ref{eq1}) means that each data point could be reconstructed by all cluster centroids and its cluster label. In addition, we enforce the clustering results of source labeled data are consistent with their initial labels $\textbf{\textit{Y}}_s^l\in{\mathbf{R}^{n_{sl}\times{C}}}$.  

\subsection{Effective Label Propagation}
\par Although the source cluster labels represent the true labels in the semi-supervised setting, it is uncertain in the target domain since no supervised information provided. To address this issue, a method of graph-based label propagation (GLP)~\cite{LP2} is introduced to guide the clustering procedure on target domain, so that their predictive cluster labels are in agreement with the true labels as accurately as possible. Specifically, we propagate the labels from the labeled source data to the unlabeled source and target data, and the loss of label propagation $\textbf{\textit{L}}_{lp}$ is defined as follows:

\begin{equation}
\begin{array}{lr}\textbf{\textit{L}}_{lp}=\sum_{i,j=1}^{n_s}\textbf{\textit{W}}_{ij}||\textbf{\textit{F}}_i-\textbf{\textit{F}}_j||_F^2+\sum_{i,j=1}^{n_s+n_t}\textbf{\textit{W}}_{ij}||\textbf{\textit{F}}_i-\textbf{\textit{F}}_j||_F^2 \\
\\
=tr(\textbf{\textit{F}}_s^{u^{\top}}\textbf{\textit{L}}_{ss}^{uu}\textbf{\textit{F}}_s^u)+2tr(\textbf{\textit{F}}_s^{u^{\top}}\textbf{\textit{L}}_{ss}^{ul}\textbf{\textit{Y}}_s^l)+tr(\textbf{\textit{F}}_t^{\top}\textbf{\textit{L}}_{tt}\textbf{\textit{F}}_t)+2tr(\textbf{\textit{F}}_t^{\top}\textbf{\textit{L}}_{ts}\textbf{\textit{F}}_s)
\quad s.t. \quad  \textbf{\textit{F}}_s^u\succeq{\textbf{\textit{0}}},\textbf{\textit{F}}_t\succeq{\textbf{\textit{0}}},
\end{array}
\label{eq2}
\end{equation}

\noindent where $\textbf{\textit{F}}_s=[\textbf{\textit{Y}}_s^l;\textbf{\textit{F}}_s^u]\in{\mathbf{R}^{n_{s}\times{C}}}$ and $\textbf{\textit{L}}=\textbf{\textit{D}}-\textbf{\textit{W}}$ represents the graph Laplacian matrix. Meanwhile, $\textbf{\textit{D}}$ denotes a diagonal matrix with the diagonal entries as the column sums of $\textbf{\textit{W}}$. Specifically,

\begin{equation}
\begin{array}{lr}
\textbf{\textit{L}}=
\begin{bmatrix}\textbf{\textit{L}}_{ss} & \textbf{\textit{L}}_{st}
\\
\\ \textbf{\textit{L}}_{ts} & \textbf{\textit{L}}_{tt}
\end{bmatrix},
\textbf{\textit{L}}_{ss}=
\begin{bmatrix}\textbf{\textit{L}}_{ss}^{ll} & \textbf{\textit{L}}_{ss}^{lu}
\\
\\ \textbf{\textit{L}}_{ss}^{ul} & \textbf{\textit{L}}_{ss}^{uu}
\end{bmatrix}.
\end{array}
\label{eq3}
\end{equation}

\subsection{Cross-Domain Feature Alignment}

\par In order to align the domain-wise distributions between the source and target domains, the MMD is adopted to explicitly reduce their marginal distribution difference, and its loss $\textbf{\textit{L}}_{mmd1}$ is defined as follows:

\begin{equation}
\begin{array}{lr}
\textbf{\textit{L}}_{mmd1}=||\frac{1}{n_s}\sum_{i=1}^{n_s}\textbf{\textit{A}}^{\top}\textbf{\textit{x}}_i-\frac{1}{n_t}\sum_{j=1}^{n_t}\textbf{\textit{A}}^{\top}\textbf{\textit{x}}_j||_2^2=tr(\textbf{\textit{A}}^{\top}\textbf{\textit{X}}\textbf{\textit{M}}_0\textbf{\textit{X}}^{\top}\textbf{\textit{A}}),
\end{array}
\label{eq4}
\end{equation}

\noindent where $\textbf{\textit{M}}_0\in{\mathbf{R}^{n\times{n}}} (n_s+n_t=n)$ is the MMD matrix, and it is computed as follows:
\begin{equation}
(\textbf{\textit{M}}_0)_{ij}=\left \{ 
\begin{array}{lr}

\frac{1}{n_sn_s},  \textbf{\textit{x}}_i,\textbf{\textit{x}}_j\in{D_s} &  \\
\\

\frac{1}{n_tn_t},  \textbf{\textit{x}}_i,\textbf{\textit{x}}_j\in{D_t} &  \\
\\

\frac{-1}{n_sn_t}, otherwise. &  

\end{array}
\right .
\label{eq5}
\end{equation}

\noindent We further decrease the conditional distribution shift across domains by class-wise MMD, and the formula $\textbf{\textit{L}}_{mmd2}$ is as follows:

\begin{equation}
\begin{array}{lr}
\textbf{\textit{L}}_{mmd2}=\sum_{c=1}^{C}||\frac{1}{n_s^c}\sum_{i=1}^{n_s^c}\textbf{\textit{A}}^{\top}\textbf{\textit{x}}_i-\frac{1}{n_t^c}\sum_{j=1}^{n_t^c}\textbf{\textit{A}}^{\top}\textbf{\textit{x}}_j||_2^2=\sum_{c=1}^{C}tr(\textbf{\textit{A}}^{\top}\textbf{\textit{X}}\textbf{\textit{M}}_c\textbf{\textit{X}}^{\top}\textbf{\textit{A}}),
\end{array}
\label{eq6}
\end{equation}

\noindent where $n_s^c$ and $n_t^c$ are the numbers of data samples from class $c$ in the source and target domains ($D_s^c$, $D_t^c$, $c\in{1,...,C}$), then the class-wise $\textbf{\textit{M}}_c$ is computed as follows:

\begin{equation}
(\textbf{\textit{M}}_c)_{ij}=\left \{ 
\begin{array}{lr}

\frac{1}{n_s^cn_s^c}, \textbf{\textit{x}}_i,\textbf{\textit{x}}_j\in{D_s^c} &  \\
\\

\frac{1}{n_t^cn_t^c}, \textbf{\textit{x}}_i,\textbf{\textit{x}}_j\in{D_t^c} &  \\
\\

-\frac{1}{n_s^cn_t^c}, \left \{ \begin{array}{lr}
\textbf{\textit{x}}_i\in{D_s^c}, \textbf{\textit{x}}_j\in{D_t^c}  \\ \\ \textbf{\textit{x}}_j\in{D_s^c}, \textbf{\textit{x}}_i\in{D_t^c} \end{array} \right.    & \\
\\

0, (otherwise). &  

\end{array}
\right.
\label{eq7}
\end{equation}

\subsection{Overall Objective Function}
\par Finally, we formulate the proposed model by incorporating the above Eq.(\ref{eq2}), Eq.(\ref{eq4}), Eq.(\ref{eq6}) as follows: 

\begin{equation}
\begin{array}{lr}
\min\limits_{\textbf{\textit{A}},\textbf{\textit{G}}_s,\textbf{\textit{F}}_s^u,\textbf{\textit{G}}_t,\textbf{\textit{F}}_t}
\gamma\textbf{\textit{L}}_{pc}+\textbf{\textit{L}}_{mmd1}+\textbf{\textit{L}}_{mmd2}+\textbf{\textit{L}}_{lp}+\lambda||\textbf{\textit{A}}||_F^2 \quad s.t. \quad \textbf{\textit{G}}_s\succeq{\textbf{\textit{0}}},\textbf{\textit{G}}_t\succeq{\textbf{\textit{0}}}, \textbf{\textit{F}}_s^u\succeq{\textbf{\textit{0}}},\textbf{\textit{F}}_t\succeq{\textbf{\textit{0}}}, \textbf{\textit{A}}^{\top}\textbf{\textit{X}}\textbf{\textit{X}}^{\top}\textbf{\textit{A}}=\textbf{\textit{I}}_m,
\end{array}
\label{eq8}
\end{equation} 

\noindent where $\gamma$, $\lambda$ are trade-off parameters, and we constrain the subspace with $\textbf{\textit{A}}^{\top}\textbf{\textit{X}}\textbf{\textit{X}}^{\top}\textbf{\textit{A}}=\textbf{\textit{I}}_m$ such that the data on the subspace are statistically uncorrelated ($\textbf{\textit{I}}_m\in{\mathbf{R}^{m\times{m}}}$ is the identity matrix and the data matrix \textbf{\textit{X}} is pre-centralized). We further impose the constraint that $||\textbf{\textit{A}}||_F^2$ is small to control the scale of $\textbf{\textit{A}}$~\cite{JGSA}.
\par Remarkably, the proposed approach joints projected clustering, label propagation and distributional alignment in a unified framework. Thus, it could benefit to each other to improve the recognition for the unlabeled data in both domains. 
\par With the projected clustering, the discriminative structures of data could be exploited effectively (i.e., the data points belonging to the same cluster could be congregated together), where only a few source labels are required. With label propagation, the cluster labels of unlabeled data are revealed correctly, either in the source or target domains. The domain-invariant features mean that the feature representations of those data instances, with the same semantic (i.e., category) from different domains, are as similar as possible, while the reasons that domain-invariant features have poor performance is that different domains follow very different distributions (i.e., domain shift). Therefore, with domain shift mitigated, the domain-invariant features could be leveraged effectively. Moreover, when they are jointly optimized, the discriminative and domain-invariant features prompt a more effective graph between the source and target domains, so that a few source labels could be propagated to the unlabeled data more accurately. Meanwhile, when more accurate labels are assigned to the unlabeled data, more effective knowledge across two domains would be transferred, and more promising projected clustering performance in both domains would be achieved. As such, those three procedures could promote to each other in a unified optimization framework and the proposed approach is more robust and effective than considering them separately.     
\par However, there exist two difficulties when Eq.(\ref{eq8}) is optimized. Firstly, the $\textbf{\textit{L}}_{mmd2}$ term contains label information, thus we have to rewrite it as a formulation, where the variable $\textbf{\textit{F}}$ is involved as the cluster centroids are optimized. As mentioned before, the source and target cluster centroids in the embedded space could be computed as $\textbf{\textit{A}}^{\top}\textbf{\textit{X}}_{(s/t)}\textbf{\textit{G}}_{(s/t)}$. Remarkably, $\textbf{\textit{L}}_{mmd2}$ is nothing less than the sum of mean distances between the source and target embedded data from the same classes. Therefore, it is easily to verify that $\textbf{\textit{L}}_{mmd2}=||\textbf{\textit{A}}^{\top}\textbf{\textit{X}}_s\textbf{\textit{G}}_s-\textbf{\textit{A}}^{\top}\textbf{\textit{X}}_t\textbf{\textit{G}}_t||_F^2$, where $\textbf{\textit{G}}_{(s/t)}=\textbf{\textit{F}}_{(s/t)}(\textbf{\textit{F}}_{(s/t)}^{\top}\textbf{\textit{F}}_{(s/t)})^{-1}$, which means that the conditional distribution alignment equals to cluster centroids calibration. Therefore, we expect that the learned cluster centroids not only enable the embedded data points more separable and discriminative, but also boost their conditional distribution alignment when the cluster centroids are optimized. 

\par Another challenge is that how to enable the form of $\textbf{\textit{L}}_{pc}$ agree with $\textbf{\textit{L}}_{mmd1}$ and $\textbf{\textit{L}}_{mmd2}$ when the shared projection is optimized. We also prove that $||\textbf{\textit{A}}^{\top}\textbf{\textit{X}}_s-\textbf{\textit{A}}^{\top}\textbf{\textit{X}}_s\textbf{\textit{G}}_s\textbf{\textit{F}}_s^{\top}||_F^2
+||\textbf{\textit{A}}^{\top}\textbf{\textit{X}}_t-\textbf{\textit{A}}^{\top}\textbf{\textit{X}}_t\textbf{\textit{G}}_t\textbf{\textit{F}}_t^{\top}||_F^2=Tr(\textbf{\textit{A}}^{\top}\textbf{\textit{S}}_w^{(s)}\textbf{\textit{A}})+Tr(\textbf{\textit{A}}^{\top}\textbf{\textit{S}}_w^{(t)}\textbf{\textit{A}})$, where $\textbf{\textit{S}}_w^{(s)}$, $\textbf{\textit{S}}_w^{(t)}$ are the intra-class scatter matrix for the source and target domains, and could be computed as previous work~\cite{UTR}. Similarly, we expect that not only the marginal and conditional distributions of source and target are aligned, but also their discriminative information could be respected when the shared projection is optimized. Therefore, the projected clustering, label propagation and distributional alignment could be optimized simultaneously, and facilitate to each other.

\noindent \textbf{Theorem 1.} The projected clustering process can be rewritten as the class scatter matrix:

\begin{equation}
\begin{array}{lr}
||\textbf{\textit{A}}^{\top}\textbf{\textit{X}}_s-\textbf{\textit{A}}^{\top}\textbf{\textit{X}}_s\textbf{\textit{G}}_s\textbf{\textit{F}}_s^{\top}||_F^2
+||\textbf{\textit{A}}^{\top}\textbf{\textit{X}}_t-\textbf{\textit{A}}^{\top}\textbf{\textit{X}}_t\textbf{\textit{G}}_t\textbf{\textit{F}}_t^{\top}||_F^2=Tr(\textbf{\textit{A}}^{\top}\textbf{\textit{S}}_w^{(s)}\textbf{\textit{A}})+Tr(\textbf{\textit{A}}^{\top}\textbf{\textit{S}}_w^{(t)}\textbf{\textit{A}}),
\end{array}
\label{eq9}
\end{equation}
\noindent where $\textbf{\textit{S}}_w^{(s)}$, $\textbf{\textit{S}}_w^{(t)}$ are the intra-class scatter matrice for the source and target domains.
\vspace{5pt}

\noindent \textbf{\textit{Proof:}} Without loss generality, we prove that $||\textbf{\textit{A}}^{\top}\textbf{\textit{X}}-\textbf{\textit{A}}^{\top}\textbf{\textit{X}}\textbf{\textit{G}}\textbf{\textit{F}}^{\top}||_F^2
=Tr(\textbf{\textit{A}}^{\top}\textbf{\textit{S}}_w\textbf{\textit{A}})$. Firstly, we denote $\textbf{\textit{M}}=[\textbf{\textit{m}}_1,\textbf{\textit{m}}_2,...,\textbf{\textit{m}}_C]$, where $\textbf{\textit{m}}_c$ represents the mean of $\textbf{\textit{x}}_i$ from class $c$. As mentioned before, $\textbf{\textit{M}}=\textbf{\textit{X}}\textbf{\textit{G}}$. Then, we have:

\begin{equation}
\begin{array}{lr}
\textbf{\textit{S}}_w=\sum_{c=1}^{C}\sum_{\textbf{\textit{x}}_i\in{D^{(c)}}}(\textbf{\textit{x}}_i-\textbf{\textit{m}}_c)(\textbf{\textit{x}}_i-\textbf{\textit{m}}_c)^T=(\textbf{\textit{X}}-\textbf{\textit{M}}\textbf{\textit{F}}^T)(\textbf{\textit{X}}-\textbf{\textit{M}}\textbf{\textit{F}}^T)^T.
\end{array}
\label{eq10}
\end{equation}

\noindent Furthermore,

\begin{equation}
\begin{array}{lr}
Tr(\textbf{\textit{A}}^T\textbf{\textit{S}}_w\textbf{\textit{A}})=Tr(\textbf{\textit{A}}^T(\textbf{\textit{X}}-\textbf{\textit{M}}\textbf{\textit{F}}^T)(\textbf{\textit{X}}-\textbf{\textit{M}}\textbf{\textit{F}}^T)^T\textbf{\textit{A}})=||\textbf{\textit{A}}^T\textbf{\textit{X}}-\textbf{\textit{A}}^T\textbf{\textit{M}}\textbf{\textit{F}}^T||_F^2=||\textbf{\textit{A}}^T\textbf{\textit{X}}-\textbf{\textit{A}}^T\textbf{\textit{X}}\textbf{\textit{G}}\textbf{\textit{F}}^T||_F^2.
\end{array}
\label{eq11}
\end{equation}

\noindent Thus, the Eq.(\ref{eq9}) is proved. 

\subsection{Optimization}
\par Here a alternative optimization strategy is constructed to solve Eq.(\ref{eq8}) as below. We first transform it into the augmented Lagrangian function by relaxing the non-negative constraint as follows:

\begin{equation}
\begin{array}{lr}
\min\limits \textbf{\textit{J}}=\gamma||\textbf{\textit{A}}^{\top}\textbf{\textit{X}}_s-\textbf{\textit{A}}^{\top}\textbf{\textit{X}}_s\textbf{\textit{G}}_s\textbf{\textit{F}}_s^{\top}||_F^2+\gamma||\textbf{\textit{A}}^{\top}\textbf{\textit{X}}_t-\textbf{\textit{A}}^{\top}\textbf{\textit{X}}_t\textbf{\textit{G}}_t\textbf{\textit{F}}_t^{\top}||_F^2+tr(\textbf{\textit{A}}^{\top}\textbf{\textit{X}}\textbf{\textit{M}}_0\textbf{\textit{X}}^{\top}\textbf{\textit{A}})+\sum_{c=1}^{C}tr(\textbf{\textit{A}}^{\top}\textbf{\textit{X}}\textbf{\textit{M}}_c\textbf{\textit{X}}^{\top}\textbf{\textit{A}})
\\
\\
+tr(\textbf{\textit{F}}_s^{u^{\top}}\textbf{\textit{L}}_{ss}^{uu}\textbf{\textit{F}}_s^u)+tr(\textbf{\textit{F}}_t^{\top}\textbf{\textit{L}}_{tt}\textbf{\textit{F}}_t)+2tr(\textbf{\textit{F}}_s^{u^{\top}}\textbf{\textit{L}}_{ss}^{ul}\textbf{\textit{Y}}_s^l)+2tr(\textbf{\textit{F}}_t^{\top}\textbf{\textit{L}}_{ts}\textbf{\textit{F}}_s)+\lambda||\textbf{\textit{A}}||_F^2+tr(\Phi_{1}\textbf{\textit{G}}_s^{\top})+tr(\Phi_{2}\textbf{\textit{G}}_t^{\top})+tr(\Phi_{3}\textbf{\textit{F}}_s^{u^{\top}})
\\
\\
+tr(\Phi_{4}\textbf{\textit{F}}_t^{\top}) \quad s.t. \quad \textbf{\textit{A}}^{\top}\textbf{\textit{X}}\textbf{\textit{X}}^{\top}\textbf{\textit{A}}=\textbf{\textit{I}}_m,
\end{array}
\label{eq12}
\end{equation}

\noindent where $\Phi_{1}$, $\Phi_{2}$, $\Phi_{3}$, $\Phi_{4}$ are the Lagrange multipliers for constraints $\textbf{\textit{G}}_s\succeq{\textbf{\textit{0}}},\textbf{\textit{G}}_t\succeq{\textbf{\textit{0}}}, \textbf{\textit{F}}_s^u\succeq{\textbf{\textit{0}}},\textbf{\textit{F}}_t\succeq{\textbf{\textit{0}}}$. When $\textbf{\textit{G}}_s,\textbf{\textit{F}}_s^u,\textbf{\textit{G}}_t,\textbf{\textit{F}}_t$ are fixed, Eq.(\ref{eq12}) becomes:

\begin{equation}
\begin{array}{lr}
\min\limits_{\textbf{\textit{A}}} tr(\textbf{\textit{A}}^{\top}\textbf{\textit{K}}_{ms}\textbf{\textit{A}}) \quad s.t. \quad \textbf{\textit{A}}^{\top}\textbf{\textit{X}}\textbf{\textit{X}}^{\top}\textbf{\textit{A}}=\textbf{\textit{I}}_m,
\end{array}
\label{eq13}
\end{equation}

\noindent where $\textbf{\textit{K}}_{ms}=\sum_{c=0}^{C}\textbf{\textit{X}}\textbf{\textit{M}}_0\textbf{\textit{X}}^{\top}+\gamma\textbf{\textit{S}}_w^{(s)}+\gamma\textbf{\textit{S}}_w^{(t)}+\lambda \textbf{\textit{I}}_m$, and $\textbf{\textit{S}}_w^{(s)}$, $\textbf{\textit{S}}_w^{(t)}$ are the intra-class scatter matrix for the source and target domains and could be computed as previous work~\cite{UTR}. Here we rewrite the cluster projection as class scatter matrix since the labels are uncovered when $\textbf{\textit{A}}$ optimized.   Then, the optimal solution $\textbf{\textit{A}}$ to Eq.(\ref{eq13}) is formed by the $k$ eigenvectors of $\textbf{\textit{K}}_{ms}$ corresponding to the $k$ smallest eigenvalues.

\noindent When $\textbf{\textit{A}},\textbf{\textit{F}}_s^u,\textbf{\textit{G}}_t,\textbf{\textit{F}}_t$ are fixed, Eq.(\ref{eq12}) becomes:

\begin{equation}
\begin{array}{lr}
\min\limits_{\textbf{\textit{G}}_s}\textbf{\textit{J}}=\min\limits_{\textbf{\textit{G}}_s}||\textbf{\textit{A}}^{\top}\textbf{\textit{X}}_s\textbf{\textit{G}}_s-\textbf{\textit{A}}^{\top}\textbf{\textit{X}}_t\textbf{\textit{G}}_t||_F^2+\gamma||\textbf{\textit{A}}^{\top}\textbf{\textit{X}}_s-\textbf{\textit{A}}^{\top}\textbf{\textit{X}}_s\textbf{\textit{G}}_s\textbf{\textit{F}}_s^{\top}||_F^2 +tr(\Phi_{1}\textbf{\textit{G}}_s^{\top}),
\end{array}
\label{eq14}
\end{equation}

\noindent where we rewrite the distributional alignment as cluster centroids calibration since the labels are unknown. Thus, we obtain the partial derivative of $\textbf{\textit{J}}$ w.r.t., $\textbf{\textit{G}}_s$, by setting it to zero as:

\begin{equation}
\begin{array}{lr}
\frac{\partial \textbf{\textit{J}}}{\partial \textbf{\textit{G}}_s}=2\textbf{\textit{X}}_s^{\top}\textbf{\textit{A}}\textbf{\textit{A}}^{\top}\textbf{\textit{X}}_s\textbf{\textit{G}}_s-2\textbf{\textit{X}}_s^{\top}\textbf{\textit{A}}\textbf{\textit{A}}^{\top}\textbf{\textit{X}}_t\textbf{\textit{G}}_t+2\gamma\textbf{\textit{X}}_s^{\top}\textbf{\textit{A}}\textbf{\textit{A}}^{\top}\textbf{\textit{X}}_s\textbf{\textit{G}}_s\textbf{\textit{F}}_s^{\top}\textbf{\textit{F}}_s-2\gamma\textbf{\textit{X}}_s^{\top}\textbf{\textit{A}}\textbf{\textit{A}}^{\top}\textbf{\textit{X}}_s\textbf{\textit{F}}_s+\Phi_{1}=\textbf{\textit{0}}.
\end{array}
\label{eq15}
\end{equation}

\noindent Using the KKT conditions $\Phi_{1}\odot \textbf{\textit{G}}_s= \textbf{\textit{0}}$ ($\odot$ denotes the dot product of two matrix), we achieve the following equations for $\textbf{\textit{G}}_s$:

\begin{equation}
\begin{array}{lr}
[2\textbf{\textit{X}}_s^{\top}\textbf{\textit{A}}\textbf{\textit{A}}^{\top}\textbf{\textit{X}}_s\textbf{\textit{G}}_s-2\textbf{\textit{X}}_s^{\top}\textbf{\textit{A}}\textbf{\textit{A}}^{\top}\textbf{\textit{X}}_t\textbf{\textit{G}}_t+2\gamma\textbf{\textit{X}}_s^{\top}\textbf{\textit{A}}\textbf{\textit{A}}^{\top}\textbf{\textit{X}}_s\textbf{\textit{G}}_s\textbf{\textit{F}}_s^{\top}\textbf{\textit{F}}_s-2\gamma\textbf{\textit{X}}_s^{\top}\textbf{\textit{A}}\textbf{\textit{A}}^{\top}\textbf{\textit{X}}_s\textbf{\textit{F}}_s]\odot \textbf{\textit{G}}_s=-\Phi_{1}\odot \textbf{\textit{G}}_s=\textbf{\textit{0}}.
\end{array}
\label{eq16}
\end{equation}

\noindent Following~\cite{GAKT, CSNMF}, we obtain the updating rule:
\begin{equation} 
\textbf{\textit{G}}_s=\textbf{\textit{G}}_s\odot\sqrt{\frac{\textbf{\textit{T}}\textbf{\textit{G}}_1+[\textbf{\textit{T}}_1]^{^-}\textbf{\textit{G}}_s+\gamma[\textbf{\textit{T}}_1]^{^-}\textbf{\textit{G}}_s\textbf{\textit{T}}_3}{\textbf{\textit{T}}\textbf{\textit{G}}_2+[\textbf{\textit{T}}_1]^{^+}\textbf{\textit{G}}_s+\gamma[\textbf{\textit{T}}_1]^{^+}\textbf{\textit{G}}_s\textbf{\textit{T}}_3}},
\label{eq17}
\end{equation}

\noindent where $T_1=\textbf{\textit{X}}_s^{\top}\textbf{\textit{A}}\textbf{\textit{A}}^{\top}\textbf{\textit{X}}_s$, $T_2=\textbf{\textit{X}}_s^{\top}\textbf{\textit{A}}\textbf{\textit{A}}^{\top}\textbf{\textit{X}}_t$, $T_3=\textbf{\textit{F}}_s^{\top}\textbf{\textit{F}}_s$, $\textbf{\textit{T}}\textbf{\textit{G}}_1=[\textbf{\textit{T}}_2]^{^+}\textbf{\textit{G}}_t+\gamma[\textbf{\textit{T}}_1]^{^+}\textbf{\textit{F}}_s$, $\textbf{\textit{T}}\textbf{\textit{G}}_2=[\textbf{\textit{T}}_2]^{^-}\textbf{\textit{G}}_t+\gamma[\textbf{\textit{T}}_1]^{^-}\textbf{\textit{F}}_s$. Moreover, $[\textbf{\textit{T}}]^{^+}$ is a matrix that the negative elements of an arbitrary matrix $\textbf{\textit{T}}$ are replaced by 0. Similarly, $[\textbf{\textit{T}}]^{^-}$ is a matrix that the positive elements of an arbitrary matrix $\textbf{\textit{T}}$ are replaced by 0. Similarly,

\begin{equation}
\textbf{\textit{G}}_t=\textbf{\textit{G}}_t\odot\sqrt{\frac{\textbf{\textit{R}}\textbf{\textit{G}}_1+[\textbf{\textit{R}}_1]^{^-}\textbf{\textit{G}}_t+\gamma[\textbf{\textit{R}}_1]^{^-}\textbf{\textit{G}}_t\textbf{\textit{R}}_3}{\textbf{\textit{R}}\textbf{\textit{G}}_2+[\textbf{\textit{R}}_1]^{^+}\textbf{\textit{G}}_t+\gamma[\textbf{\textit{R}}_1]^{^+}\textbf{\textit{G}}_t\textbf{\textit{R}}_3}},
\label{eq18}
\end{equation}

\noindent where $\textbf{\textit{R}}_1=\textbf{\textit{X}}_t^{\top}\textbf{\textit{A}}\textbf{\textit{A}}^{\top}\textbf{\textit{X}}_t$, $\textbf{\textit{R}}_2=\textbf{\textit{X}}_t^{\top}\textbf{\textit{A}}\textbf{\textit{A}}^{\top}\textbf{\textit{X}}_s$, $\textbf{\textit{R}}_3=\textbf{\textit{F}}_t^{\top}\textbf{\textit{F}}_t$, $\textbf{\textit{R}}\textbf{\textit{G}}_1=[\textbf{\textit{R}}_2]^{^+}\textbf{\textit{G}}_s+\gamma[\textbf{\textit{R}}_1]^{^+}\textbf{\textit{F}}_t$, $\textbf{\textit{R}}\textbf{\textit{G}}_2=[\textbf{\textit{R}}_2]^{^-}\textbf{\textit{G}}_s+\gamma[\textbf{\textit{R}}_1]^{^-}\textbf{\textit{F}}_t$.

\par As for $\textbf{\textit{F}}_s^{u}$, we fix $\textbf{\textit{A}},\textbf{\textit{G}}_s,\textbf{\textit{G}}_t,\textbf{\textit{F}}_t$ and Eq.(\ref{eq12}) becomes:
\begin{equation}
\begin{array}{lr}
\min\limits_{\textbf{\textit{F}}_s^u} \textbf{\textit{J}}
=\min\limits_{\textbf{\textit{F}}_s^u}
\gamma||\textbf{\textit{A}}^{\top}\textbf{\textit{X}}_s^u-\textbf{\textit{A}}^{\top}\textbf{\textit{X}}_s\textbf{\textit{G}}_s\textbf{\textit{F}}_s^{u^{\top}}||_F^2+tr(\textbf{\textit{F}}_s^{u^{\top}}\textbf{\textit{L}}_{ss}^{uu}\textbf{\textit{F}}_s^u)+2tr(\textbf{\textit{F}}_s^{u^{\top}}\textbf{\textit{L}}_{ss}^{ul}\textbf{\textit{Y}}_s^l)+tr(\Phi_{3}\textbf{\textit{F}}_s^{u^{\top}}).
\end{array}
\label{eq19}
\end{equation}

\noindent Likewise, we obtain the following equations for $\textbf{\textit{F}}_s^u$:

\begin{equation}
\setlength\abovedisplayskip{1pt plus 3pt minus 7pt} 
\setlength\belowdisplayskip{10pt plus 3pt minus 7pt} 
\begin{array}{lr}
[2\textbf{\textit{L}}_{ss}^{uu}\textbf{\textit{F}}_s^u+2\textbf{\textit{L}}_{ss}^{ul}\textbf{\textit{Y}}_s^l+2\gamma\textbf{\textit{F}}_s^u\textbf{\textit{G}}_s^{\top}\textbf{\textit{X}}_s^{\top}\textbf{\textit{A}}\textbf{\textit{A}}^{\top}\textbf{\textit{X}}_s\textbf{\textit{G}}_s-2\gamma\textbf{\textit{X}}_s^{u^{\top}}\textbf{\textit{A}}\textbf{\textit{A}}^{\top}\textbf{\textit{X}}_s\textbf{\textit{G}}_s]\odot \textbf{\textit{F}}_s^u=-\Phi_{3}\odot \textbf{\textit{F}}_s^u=\textbf{\textit{0}}.
\end{array}
\label{eq20}
\end{equation}

\noindent Therefore, the updating rule for $\textbf{\textit{F}}_s^u$ is as follows:

\begin{equation}
\textbf{\textit{F}}_s^u=\textbf{\textit{F}}_s^u\odot\sqrt{\frac{\gamma[\textbf{\textit{K}}_1]^{^+}+\gamma\textbf{\textit{F}}_s^u[\textbf{\textit{K}}_2]^{^-}+[\textbf{\textit{L}}_{ss}^{uu}]^{^-}\textbf{\textit{F}}_s^u+[\textbf{\textit{L}}_{ss}^{ul}]^{^-}\textbf{\textit{Y}}_s^l}{\gamma[\textbf{\textit{K}}_1]^{^-}+\gamma\textbf{\textit{F}}_s^u[\textbf{\textit{K}}_2]^{^+}+[\textbf{\textit{L}}_{ss}^{uu}]^{^+}\textbf{\textit{F}}_s^u+[\textbf{\textit{L}}_{ss}^{ul}]^{^+}\textbf{\textit{Y}}_s^l}},
\label{eq21}
\end{equation}

\noindent where $\textbf{\textit{K}}_1=\textbf{\textit{X}}_s^{u^{\top}}\textbf{\textit{A}}\textbf{\textit{A}}^{\top}\textbf{\textit{X}}_s\textbf{\textit{G}}_s$, $\textbf{\textit{K}}_2=\textbf{\textit{G}}_s^{\top}\textbf{\textit{X}}_s^{\top}\textbf{\textit{A}}\textbf{\textit{A}}^{\top}\textbf{\textit{X}}_s\textbf{\textit{G}}_s$. Similarly, the updating rule for $\textbf{\textit{F}}_t$ is as follows:

\begin{equation}
\textbf{\textit{F}}_t=\textbf{\textit{F}}_t\odot\sqrt{\frac{\gamma[\textbf{\textit{K}}_3]^{^+}+\gamma\textbf{\textit{F}}_t[\textbf{\textit{K}}_4]^{^-}+[\textbf{\textit{L}}_{tt}]^{^-}\textbf{\textit{F}}_t+[\textbf{\textit{L}}_{ts}]^{^-}\textbf{\textit{F}}_s}{\gamma[\textbf{\textit{K}}_3]^{^+}+\gamma\textbf{\textit{F}}_t[\textbf{\textit{K}}_4]^{^-}+[\textbf{\textit{L}}_{tt}]^{^-}\textbf{\textit{F}}_t+[\textbf{\textit{L}}_{ts}]^{^-}\textbf{\textit{F}}_s}},
\label{eq22}
\end{equation}

\noindent where $\textbf{\textit{K}}_3=\textbf{\textit{X}}_t^{\top}\textbf{\textit{A}}\textbf{\textit{A}}^{\top}\textbf{\textit{X}}_t\textbf{\textit{G}}_t$, $\textbf{\textit{K}}_4=\textbf{\textit{G}}_t^{\top}\textbf{\textit{X}}_t^{\top}\textbf{\textit{A}}\textbf{\textit{A}}^{\top}\textbf{\textit{X}}_t\textbf{\textit{G}}_t$.

\par To make these update rules clear, we summarize the algorithm to solve Eq.(\ref{eq12}) in~Algorithm \ref{Alg1}.

\begin{algorithm}
	\caption{\textbf{SLSA-DA}}
	\KwIn{Sparsely labeled source data and the limited source labels, $\textbf{\textit{X}}_s=[\textbf{\textit{X}}_s^l,\textbf{\textit{X}}_s^u]$, $\textbf{\textit{Y}}_s^l$.\\
	\qquad \qquad Unlabeled target data $\textbf{\textit{X}}_t$. Regularized parameters $\textit{k}$, $\gamma$, $\lambda$, $T$}
	\KwOut{Labels of source and target unlabeled data (i.e., $\textbf{\textit{Y}}_s^u$ and $\textbf{\textit{Y}}_t$) for $\textbf{\textit{X}}_s^u$ and $\textbf{\textit{X}}_t$}
	\textbf{Begin} \\
	\textbf{Initialization} \\
	\textbf{Line 1:} Compute $\textbf{\textit{M}}_0$ by Eq.(\ref{eq5}) \\
	\textbf{Line 2:} Obtain $\textbf{\textit{A}}$ by Eq.(\ref{eq13}) with $\gamma=0$ and $C=0$ \\
	\textbf{Line 3:} Obtain $\textbf{\textit{Z}}_s=\textbf{\textit{A}}^{\top}\textbf{\textit{X}}_s$, $\textbf{\textit{Z}}_t=\textbf{\textit{A}}^{\top}\textbf{\textit{X}}_t$ \\
	\textbf{Line 4:} Propagate labels from $\textbf{\textit{Z}}_s^l$ to $\textbf{\textit{Z}}_s^u$, i.e., $\textbf{\textit{Y}}_s^u$ \\
	\textbf{Line 5:} Propagate labels from $[\textbf{\textit{Z}}_s^l,\textbf{\textit{Z}}_s^u]$ to $\textbf{\textit{Z}}_t$, i.e., $\textbf{\textit{Y}}_t$ \\
	
	\textbf{For} $t$=1 to $T$ \textbf{do} \\
	\textbf{Line 6:} Compute $\textbf{\textit{S}}_w^{(s)}$, $\textbf{\textit{S}}_w^{(t)}$, $\textbf{\textit{M}}_c,c=1,...,C$ by Eq.(\ref{eq10}) and Eq.(\ref{eq6})\\
	\textbf{Line 7:} Update $\textbf{\textit{Z}}_s$ and $\textbf{\textit{Z}}_t$ by Eq.(\ref{eq13}) \\
	\textbf{Line 8:} Update $\textbf{\textit{G}}_s$ and $\textbf{\textit{G}}_t$ by Eq.(\ref{eq17}) and Eq.(\ref{eq18}) \\
	\textbf{Line 9:} Update $\textbf{\textit{F}}_s$ and $\textbf{\textit{F}}_t$ by Eq.(\ref{eq21}) and Eq.(\ref{eq22}) \\
	\textbf{End repeat}  \\
	\textbf{Return} One-hot labels $[\textbf{\textit{Y}}_s^u;\textbf{\textit{Y}}_t]$ 
	\label{Alg1}
\end{algorithm}

\textbf{Computational Complexity}
\par We analyze the computational complexity of~Algorithm \ref{Alg1} using the $O$ notation. We denote $T$ as the number of iterations. The computational cost is detailed as follows: $O(Tkm^2)$ for solving the generalized eigen-decomposition problem, i.e. \textbf{Line 7}; $O(TCn^2)$ for updating $\textbf{\textit{G}}_s$, $\textbf{\textit{G}}_t$, i.e. \textbf{Line 8}; $O(TCn^2)$ for constructing the $\textbf{\textit{M}}_c$, i.e. \textbf{Lines 6}; $O(TCn^2+TC^2n)$ for updating $\textbf{\textit{F}}_s$, $\textbf{\textit{F}}_t$, i.e. \textbf{Line 9}; $O(Tm^2n+TCmn)$ for updating $\textbf{\textit{S}}_w^{(s)}$, $\textbf{\textit{S}}_w^{(t)}$, i.e. \textbf{Line 6}; In summary, the overall computational complexity of~Algorithm \ref{Alg1} is $O(Tkm^2+TCn^2+TC^2n+Tm^2n+TCmn)$. Moreover, the value of \textit{k} is not greater than 200, $T$ not greater than 100, so $k,T\ll min(m,n)$. Therefore, it can be solved in polynomial time with respect to the number of samples.

\section{Experiments}

\subsection{Datasets and Experimental Settings} 

\par
\begin{figure}[htp]
	\begin{center}
		\includegraphics[width=1.0\linewidth,height=0.35\textheight]{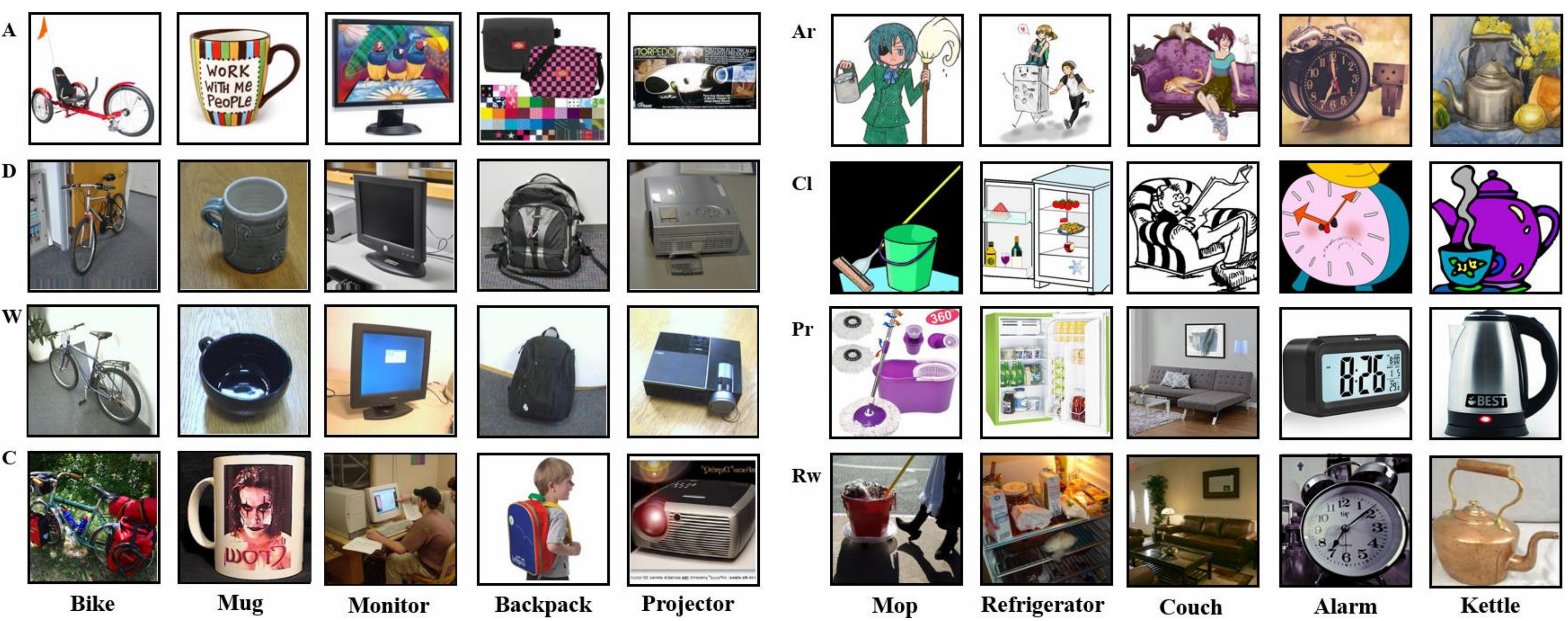}
	\end{center}
	\caption{Exemplars from (A) Amazon, (D) Dslr, (W) Webcam, (C) Caltech,  (Ar) Art, (Cl) Clipart, (Pr) Product and (Rw) Real-World datasets}
	\label{fig3}
\end{figure}

 In order to validate the effectiveness of our approach in both the DA and SLSA-DA scenario, we conducted experiments on 4 benchmark datasets in cross-domain object recognition, i.e., Office10-Caltech10, Office-Home, ImageCLEF-DA, Office31. Fig.~\ref{fig3} illustrates some sample images from Office10-Caltech10 and Office-Home datasets, and they follow very different distributions. Their descriptions are introduced as follows: 

\par \textbf{Office10-Caltech10}~\cite{GFK} contains 4 real-world object domains, where 3 domains are come from Office31 dataset (i.e., Amazon (A), Webcam (W) and DSLR (D)), and the last one is come from Caltech256 dataset (Caltech (C)). Then we select 10 shared classes between these 4 domains and construct a DA dataset Office10-Caltech10, which has 2,533 images and $4\times 3=12$ DA tasks, e.g., A$\rightarrow$W, C$\rightarrow$D and so on. Note that the arrow "$\rightarrow$" is the direction from the source domain to target domain. For example, W$\rightarrow$D means Webcam is the labeled source domain while Dslr is the unlabeled target domain.
 
\par \textbf{Office-Home}~\cite{DHN} was released recently as a more challenging dataset, crawled through several search engines and online image directories. It consists of 4 different domains, Artistic images (Ar), Clipart images (Cl), Product images (Pr) and Real-World images (Rw). In total, there are 15,500 images from 65 object categories, and 12 DA tasks.

\par \textbf{ImageCLEF-DA}~\cite{CDAN} has 1800 images organized by selecting the 12 common classes shared by 3 public domains, Caltech-256 (C), ImageNet ILSVRC 2012 (I), and Pascal VOC 2012 (P), where 6 DA tasks can be created. 

\par \textbf{Office31}~\cite{Office31} is an increasingly popular benchmark for visual DA, which includes 3 real-world object domains, Amazon (A), Webcam (W) and DSLR (D), and has 4,652 images from 31 categories, then 6 DA tasks can be constructed. 

\subsection{Experimental Results}

\begin{table}[h]
	\begin{center}
		\fontsize{5}{6}\selectfont
		\caption{Accuracy (\%) on the Office10-Caltech10 dataset with SURF features in DA setting} \label{tab1}
		\begin{tabular}{|c|c|c|c|c|c|c|c|c|c|c|c|c|c|}
			\hline
			DA &C$\rightarrow$A&C$\rightarrow$W&C$\rightarrow$D&A$\rightarrow$C&A$\rightarrow$W&A$\rightarrow$D&W$\rightarrow$C&W$\rightarrow$A&W$\rightarrow$D&D$\rightarrow$C&D$\rightarrow$A&D$\rightarrow$W&Avg.
			\\
			\hline
			\hline
			TCA~\cite{TCA}&46.5&35.3&44.6&38.6&38.3&38.9&30.2&28.7&90.4&33.3&33.4&88.8&45.6 \\
			\hline
			JDA~\cite{JDA}&43.1&35.9&47.8&34.9&42.0&36.3&31.0&40.4&88.5&29.2&29.7&89.5&45.7 \\
			\hline
			BDA~\cite{BDA}&45.4&39.7&43.9&38.6&44.7&40.8&30.1&35.0&89.8&29.9&35.9&\textbf{89.8}&47.0 \\
			\hline
			VDA~\cite{VDA}&50.8&44.7&49.0&37.5&44.7&43.9&30.9&41.3&88.5&30.2&34.1&89.5&48.8 \\
			\hline
			JGSA~\cite{JGSA}&50.7&47.5&43.9&42.5&48.1&46.5&30.9&39.9&89.8&30.0&39.0&89.2&49.8 \\
			\hline
			MEDA~\cite{MEDA}&55.4&54.6&\textbf{57.3}&44.4&\textbf{55.3}&40.1&34.3&\textbf{41.8}&86.6&33.6&43.1&86.4&52.7 \\
			\hline
			Our           &\textbf{59.2}&\textbf{58.0}&55.4&\textbf{46.4}&45.1&\textbf{47.1}&\textbf{36.5}&32.0&\textbf{93.6}&\textbf{38.3}&\textbf{44.8}&89.5&\textbf{53.8} \\
			\hline
		\end{tabular}
	\end{center}
\end{table}

\begin{table}[h]
	\vspace{-10pt}
	\begin{center}
		\fontsize{5}{6}\selectfont
		\caption{Accuracy (\%) on the ImageCLEFF-DA and Office31 datasets with ResNet50 features in DA setting}\label{tab2}
		\begin{tabular}{|c|c|c|c|c|c|c|c|c|c|c|c|c|c|}
			\hline
			DA &I$\rightarrow$P&P$\rightarrow$I&I$\rightarrow$C&C$\rightarrow$I&C$\rightarrow$P&P$\rightarrow$C&A$\rightarrow$W&D$\rightarrow$W&W$\rightarrow$D&A$\rightarrow$D&D$\rightarrow$A&W$\rightarrow$A&Avg.
			\\
			\hline
			\hline
			JAN~\cite{JAN}&76.8&88.0&94.7&89.5&74.2&91.7&85.4&97.4&99.8&84.7&68.6&70.0&85.1 \\	
			\hline
			CDAN ~\cite{CDAN}&76.7&90.6&\textbf{97.0}&90.5&74.5&93.5&\textbf{93.1}&98.2&\textbf{100.0}&\textbf{89.8}&70.1&68.0&86.8 \\
			\hline
			CAN~\cite{CAN}&78.2&87.5&94.2&89.5&75.8&89.2&81.5&98.2&99.7&85.5&65.9&63.4&84.1 \\
			\hline
			MADA~\cite{MADA}&75.0&87.9&96.0&88.8&75.2&92.2&90.1&97.4&99.6&87.8&70.3&66.4&85.6 \\
			\hline
			TCA~\cite{TCA}&77.7&81.2&92.7&87.5&74.2&84.8&76.1&97.6&99.4&79.7&64.2&63.8&81.6 \\
			\hline
			JDA~\cite{JDA}&77.0& 	81.3& 	95.2& 	91.2& 	76.8& 	84.3& 	83.3& 	98.0& 	99.8& 	81.7& 	68.2& 	69.0&	83.8 \\
			\hline
			BDA~\cite{BDA}&76.0& 	79.7& 	94.8& 	91.5& 	76.2& 	82.2& 	80.8& 	96.4& 	99.6& 	79.9& 	67.6& 	67.2& 	82.7 \\
			\hline
			VDA~\cite{VDA}&77.3& 	83.3& 	94.3& 	91.5& 	77.0& 	87.2& 	84.3& 	98.6& 	\textbf{100.0}& 	82.5& 	68.7& 	69.8& 	84.5  \\
			\hline
			JGSA~\cite{JGSA}&77.0& 	83.5& 	95.5& 	91.7& 	77.3& 	88.8& 	86.7& 	97.9& 	99.8& 	83.9& 	69.6& 	71.3& 	85.3 \\
			\hline
			MEDA~\cite{MEDA}&\textbf{79.5}& 	\textbf{92.2}& 	95.7& 	\textbf{92.3}& 	\textbf{78.7}& 	\textbf{95.5}& 	86.2& 	97.7& 	99.6& 	86.1& 	72.6& 	\textbf{74.7}& 	\textbf{87.6} \\
			\hline
			Our           &79.2& 	90.0& 	94.8& 	91.5& 	78.3& 	93.8& 	86.0& 	\textbf{98.6}& 	99.8& 	88.4& 	\textbf{74.5}& 	71.9& 	87.2 \\
			\hline
		\end{tabular}
	\end{center}
\end{table}

	\begin{table}[!h]
	\vspace{-10pt}
	\begin{center}
		\fontsize{5}{6}\selectfont
		\caption{Accuracy (\%) on the Office-Home dataset with ResNet50 features in DA setting} \label{tab3}
		\begin{tabular}{|c|c|c|c|c|c|c|c|c|c|c|c|c|c|}
			\hline
			DA
			&Ar$\rightarrow$Cl&Ar$\rightarrow$Pr&Ar$\rightarrow$Rw&Cl$\rightarrow$Ar&Cl$\rightarrow$Pr&Cl$\rightarrow$Rw&Pr$\rightarrow$Ar&Pr$\rightarrow$Cl&Pr$\rightarrow$Rw&Rw$\rightarrow$Ar&Rw$\rightarrow$Cl&Rw$\rightarrow$Pr&Avg.
			\\
			\hline
			\hline
			JAN~\cite{JAN}&45.9&61.2&68.9&50.4&59.7&61.0&45.8&43.4&70.3&63.9&52.4&76.8&58.3 \\
			\hline
			CDAN~\cite{CDAN}&50.7&70.6&76.0&57.6&70.0&70.0&57.4&50.9&77.3&70.9&56.7&81.6&65.8 \\
			\hline
			MDD~\cite{MDD}&54.9&73.7&77.8&60.0&71.4&71.8&61.2&53.6&78.1&\textbf{72.5}&\textbf{60.2}&82.3&68.1 \\
			\hline
			TADA~\cite{TADA}&53.1&72.3&77.2&59.1&71.2&72.1&59.7&53.1&78.4&72.4&60.0&\textbf{82.9}&67.6 \\
			\hline
			BSP~\cite{BSP}&52.0&68.6&76.1&58.0&70.3&70.2&58.6&50.2&77.6&72.2&59.3&81.9&66.3 \\
			\hline
			TAT~\cite{TAT}&51.6&69.5&75.4&59.4&69.5&68.6&59.5&50.5&76.8&70.9&56.6&81.6&65.8 \\
			\hline
			TCA~\cite{TCA}&48.7&65.3&70.1&49.2&59.7&63.2&52.0&45.0&71.9&63.7&51.4&77.1&59.8  \\
			\hline
			JDA~\cite{JDA}&50.9&67.7&70.9&51.3&64.4&64.9&54.6&47.7&73.3&64.9&53.7&78.3&61.9  \\
			\hline
			BDA~\cite{BDA}&47.8&59.3&67.7&49.0&62.0&61.4&50.1&46.0&70.7&61.8&51.5&74.5&58.5 \\
			\hline
			VDA~\cite{VDA}&51.2&69.3&72.2&53.6&66.1&66.9&56.0&48.8&74.5&65.8&54.1&79.5&63.2  \\
			\hline
			JGSA~\cite{JGSA}&51.4&69.2&72.6&51.8&67.3&67.0&55.9&48.7&75.6&64.4&53.3&78.5&63.0 \\
			\hline
			MEDA~\cite{MEDA}&55.3&75.7&77.6&57.2&\textbf{73.9}&72.0&58.6&52.3&78.7&68.3&57.0&81.9&67.4 \\
			\hline
			Our           &\textbf{58.1}&\textbf{77.4}&\textbf{78.7}&\textbf{61.6}&72.5&\textbf{72.5}&\textbf{62.5}&\textbf{54.4}&\textbf{79.1}&70.1&59.6&82.6&\textbf{69.1} \\
			\hline
		\end{tabular}
	\end{center}
\end{table}

\par The proposed approach involves 4 parameters: projected clustering regularizer $\gamma$, projected scaling regularizer $\lambda$, subspace dimensions $k$ and iterations $T$. For the parameters, we fix $T$=5, $\gamma$=0.01, and the 20-nearest neighbor graph is adopted with Euclidean distance-based weight for simplicity. Specially, we set $k$=20, $\lambda$=0.05 on the Office10-Caltech10 and ImageCLEFF-DA datasets, while $k$=100, $\lambda$=0.1 on the Office-Home and Office31 datasets since they contain more categories. In the coming section, we will provide empirical analysis on parameter sensitivity, which verifies that a stable performance could be achieved under a wide range of values. 

\par We adopted different types of features as the inputs, either the traditional shallow features or deep features. Specifically, the shallow SURF features~\cite{GFK} with 800 dimensions are adopted in Office10-Caltech10. As for Office-Home, ImageCLEF-DA and Office31, we utilize the deep features pre-extracted from the ResNet50 model and pre-trained on ImageNet~\cite{ResNet}, and the feature dimensionality is 2048. In order to construct an SLSA-DA scenario, we randomly choose 5 source instances from each class as labeled samples and others are unlabeled, then the random selection is repeated ten times and average results are adopted.

\par Since no previous approaches have been proposed to tackle SLSA-DA problem, we first compare the proposed approach with several state-of-art methods in DA setting, where the labels for all of source data instances are available. Specifically, in DA setting, we compare the proposed approach with both the shallow DA methods (TCA~\cite{TCA}, JDA~\cite{JDA}, BDA~\cite{BDA}, VDA~\cite{VDA}, JGSA~\cite{JGSA}, MEDA~\cite{MEDA}) and the deep DA methods (JAN~\cite{JAN}, CAN~\cite{CAN}, MADA~\cite{MADA}, CDAN~\cite{CDAN}, MDD~\cite{MDD}, TADA~\cite{TADA}, BSP~\cite{BSP}, TAT~\cite{TAT}). The performances of different methods in DA settings are shown in Table~\ref{tab1}, Table~\ref{tab2}, Table~\ref{tab3},To be specific, Table~\ref{tab1} illustrates that the results of our approach are substantially higher than all other 6 ones on most DA tasks (8/12), and the average accuracy is 53.8\%, which has 1.1\% improvement compared with the best baseline MEDA. From Table~\ref{tab2}, it could be seen that best results are achieved only 2/12 DA tasks but most of them are very close to the highest ones. Besides, the average accuracy of our approach is only 0.2\% lower than the best baseline MEDA. From Table~\ref{tab3}, it can be observed that our approach is also able to attain best performances on the most DA tasks (8/12), and increase the average accuracy by 1.0\% compared with the best baseline MDD (68.1\% to 69.1\%). Therefore, the competitive capability of our approach in DA setting could be validated compared with those state-of-art DA methods, either the shallow or deep ones.


\par In order to further embody the superiority of our approach concerning the SLSA-DA scenario, we also test the behavior of other mainstream approaches in the SLSA-DA problem. As a note, the deep DA methods integrate feature extraction and knowledge transfer into an end-to-end network and achieve promising results, and this paper adopts a two-stage mechanism to promote the transferability of deep ResNet50 features. Some recent techniques have been proven that more effective knowledge transfer, is easier and faster to be implemented with this two-stage mechanism. Moreover, the promising results of deep DA methods mainly depend on feeding adequate labeled data, it may well fail to train a classifier since the labels are very limited in SLSA-DA scenario. Therefore, in SLSA-DA scenario, we only report the results compared with those two-stage methods (i.e., TCA~\cite{TCA}, JDA~\cite{JDA}, BDA~\cite{BDA}, VDA~\cite{VDA}, JGSA~\cite{JGSA}, MEDA~\cite{MEDA}). 

\begin{table}[!h]
	\begin{center}
		\fontsize{5}{6}\selectfont
		\caption{Accuracy (\%) on the Office10-Caltech10 dataset with SURF features in SLSA-DA setting} \label{tab4}
		\begin{tabular}{|c|c|c|c|c|c|c|c|c|c|c|c|c|c|}
			\hline
			SLSA-DA &C$\rightarrow$A&C$\rightarrow$W&C$\rightarrow$D&A$\rightarrow$C&A$\rightarrow$W&A$\rightarrow$D&W$\rightarrow$C&W$\rightarrow$A&W$\rightarrow$D&D$\rightarrow$C&D$\rightarrow$A&D$\rightarrow$W&Avg.
			\\
			\hline
			\hline
			TCA(s)~\cite{TCA}&39.5$\pm$3.1&36.5$\pm$2.3&38.1$\pm$2.9&53.1$\pm$2.6&54.7$\pm$2.6&54.0$\pm$3.1&79.4$\pm$2.4&75.8$\pm$4.3&80.5$\pm$4.0&\textbf{79.5}$\pm$1.7&75.7$\pm$2.7&81.8$\pm$3.9&62.4$\pm$3.0  \\
			\hline
			JDA(s)~\cite{JDA}&37.5$\pm$3.0&33.8$\pm$2.3&35.1$\pm$2.8&47.1$\pm$3.6&47.1$\pm$2.9&47.9$\pm$2.2&68.9$\pm$4.4&67.3$\pm$4.4&76.9$\pm$4.1&73.4$\pm$3.1&70.1$\pm$2.8&81.7$\pm$2.9&57.2$\pm$3.2  \\
			\hline
			BDA(s)~\cite{BDA}&39.7$\pm$2.3&35.9$\pm$2.9&38.0$\pm$2.6&49.4$\pm$2.5&52.5$\pm$2.6&51.4$\pm$2.7&73.8$\pm$3.6&71.5$\pm$4.0&79.7$\pm$4.3&74.3$\pm$2.5&72.2$\pm$1.7&81.7$\pm$2.2&60.0$\pm$2.8  \\
			\hline
			VDA(s)~\cite{VDA}&39.1$\pm$2.7&34.5$\pm$2.5&36.1$\pm$3.1&47.8$\pm$3.6&49.2$\pm$2.3&49.7$\pm$2.7&68.2$\pm$3.7&65.1$\pm$4.3&78.6$\pm$3.3&72.9$\pm$3.2&68.2$\pm$4.0&80.9$\pm$3.2&57.5$\pm$3.2  \\  
			\hline
		  JGSA(s)~\cite{JGSA}&41.2$\pm$2.6&33.1$\pm$2.1&32.8$\pm$2.3&50.4$\pm$3.1&45.7$\pm$3.6&43.6$\pm$4.8&71.0$\pm$3.3&66.5$\pm$4.5&79.7$\pm$2.1&75.0$\pm$3.4&72.7$\pm$2.2&80.3$\pm$4.4&57.7$\pm$3.2  \\
			\hline
		  MEDA(s)~\cite{MEDA}&39.5$\pm$3.8&35.2$\pm$2.7&35.5$\pm$2.4&53.6$\pm$3.1&53.0$\pm$3.0&50.4$\pm$4.1&77.1$\pm$2.9&77.3$\pm$3.0&78.1$\pm$3.3&77.4$\pm$2.8&\textbf{76.8}$\pm$3.5&78.3$\pm$4.4&61.0$\pm$3.3  \\
		    \hline
			Our(s)           &\textbf{45.1}$\pm$2.5&\textbf{40.9}$\pm$3.1&\textbf{42.4}$\pm$1.8&\textbf{58.4}$\pm$3.4&\textbf{60.3}$\pm$2.5&\textbf{59.4}$\pm$1.9&\textbf{80.0}$\pm$3.5&\textbf{79.4}$\pm$2.6&\textbf{85.1}$\pm$3.4&77.3$\pm$3.0&76.5$\pm$2.2&\textbf{86.1}$\pm$3.0&\textbf{65.9}$\pm$2.7  \\
			\hline
			
		   TCA(t)~\cite{TCA} &33.9$\pm$3.5&27.8$\pm$6.1&33.5$\pm$4.1&31.5$\pm$1.5&30.2$\pm$2.7&30.9$\pm$2.3&27.9$\pm$2.6&28.0$\pm$1.3&73.4$\pm$5.8&31.4$\pm$1.4&32.6$\pm$1.9&79.6$\pm$2.8&38.4$\pm$3.0  \\
		   \hline
		   JDA(t)~\cite{JDA} &33.4$\pm$3.7&28.0$\pm$5.7&31.5$\pm$4.4&28.8$\pm$3.0&32.1$\pm$2.2&31.2$\pm$3.4&28.2$\pm$3.0&28.3$\pm$6.4&69.4$\pm$6.5&29.8$\pm$2.6&31.7$\pm$2.6&79.3$\pm$2.7&37.6$\pm$3.9  \\
		   \hline
		   BDA(t)~\cite{BDA} &34.4$\pm$3.4&30.7$\pm$4.9&34.5$\pm$5.5&31.4$\pm$2.0&33.9$\pm$4.6&33.5$\pm$2.9&31.0$\pm$2.1&30.4$\pm$4.7&71.8$\pm$5.6&29.8$\pm$1.8&32.6$\pm$2.7&80.6$\pm$2.5&39.6$\pm$3.6  \\
		   \hline
		   VDA(t)~\cite{VDA} &37.0$\pm$4.7&30.7$\pm$5.7&33.1$\pm$5.0&30.6$\pm$2.9&33.9$\pm$3.0&30.6$\pm$4.0&29.3$\pm$2.8&28.8$\pm$5.8&73.2$\pm$6.2&30.7$\pm$2.6&34.8$\pm$2.8&78.9$\pm$2.5&39.3$\pm$4.0  \\
	       \hline
		 JGSA(t)~\cite{JGSA} &38.4$\pm$4.7&32.3$\pm$7.9&34.2$\pm$3.4&33.5$\pm$2.8&36.1$\pm$8.3&33.1$\pm$4.2&28.7$\pm$1.5&36.0$\pm$3.1&73.8$\pm$2.5&29.0$\pm$1.2&34.9$\pm$3.4&79.1$\pm$4.3&40.8$\pm$3.9  \\
		 \hline
		 MEDA(t)~\cite{MEDA} &38.5$\pm$5.4&34.8$\pm$5.1&37.7$\pm$5.0&35.6$\pm$2.5&\textbf{38.4}$\pm$6.4&33.6$\pm$4.2&30.6$\pm$3.1&\textbf{38.1}$\pm$2.6&72.0$\pm$5.5&35.2$\pm$2.0&35.2$\pm$1.8&74.7$\pm$4.2&42.0$\pm$4.0  \\
		 \hline
		   Our(t)            &\textbf{42.2}$\pm$3.6&\textbf{37.3}$\pm$7.1&\textbf{39.2}$\pm$4.9&\textbf{38.8}$\pm$1.7&37.0$\pm$6.7&\textbf{38.9}$\pm$2.6&\textbf{33.4}$\pm$1.7&32.4$\pm$2.2&\textbf{78.3}$\pm$3.4&\textbf{36.1}$\pm$3.2&\textbf{40.1}$\pm$3.7&\textbf{82.2}$\pm$3.8&\textbf{44.7}$\pm$3.7  \\
		   \hline
		\end{tabular}
	\end{center}
\end{table}

\begin{table}[!h]
	\begin{center}
		\fontsize{5}{6}\selectfont
		\caption{Accuracy (\%) on the ImageCLEFF-DA and Office31 datasets with ResNet50 features in SLSA-DA setting} \label{tab5}
		\begin{tabular}{|c|c|c|c|c|c|c|c|c|c|c|c|c|c|}
			\hline
			SLSA-DA &I$\rightarrow$P&P$\rightarrow$I&I$\rightarrow$C&C$\rightarrow$I&C$\rightarrow$P&P$\rightarrow$C&A$\rightarrow$W&D$\rightarrow$W&W$\rightarrow$D&A$\rightarrow$D&D$\rightarrow$A&W$\rightarrow$A&Avg.
			\\
			\hline
			\hline
			TCA(s)~\cite{TCA} &94.2$\pm$0.8&71.7$\pm$2.7&93.7$\pm$0.9&97.4$\pm$0.4&97.3$\pm$0.6&71.0$\pm$3.1&86.7$\pm$0.8&97.7$\pm$0.8&96.5$\pm$1.2&86.2$\pm$0.7&\textbf{98.0}$\pm$0.5&96.7$\pm$0.9&90.6$\pm$1.1  \\
			\hline
			JDA(s)~\cite{JDA} &91.2$\pm$1.0&70.7$\pm$3.5&94.1$\pm$0.7&95.5$\pm$1.4&93.5$\pm$1.2&72.2$\pm$3.5&85.5$\pm$1.0&97.0$\pm$1.0&96.4$\pm$1.4&85.3$\pm$0.6&95.5$\pm$0.5&92.9$\pm$1.1&89.2$\pm$1.4  \\
			\hline
			BDA(s)~\cite{BDA} &87.6$\pm$1.4&70.5$\pm$3.5&93.7$\pm$0.9&96.0$\pm$1.3&93.7$\pm$1.3&71.2$\pm$3.7&83.2$\pm$1.0&96.3$\pm$1.0&95.6$\pm$1.2&83.9$\pm$1.1&92.1$\pm$2.0&91.9$\pm$1.5&88.0$\pm$1.7  \\
			\hline
			VDA(s)~\cite{VDA} &93.6$\pm$0.6&72.8$\pm$3.6&95.1$\pm$0.8&96.9$\pm$1.0&95.7$\pm$0.7&74.0$\pm$3.6&86.5$\pm$1.0&97.6$\pm$0.9&96.6$\pm$1.2&86.1$\pm$0.8&96.1$\pm$0.8&93.3$\pm$1.0&90.4$\pm$1.3  \\
			\hline
		  JGSA(s)~\cite{JGSA} &94.3$\pm$0.6&72.5$\pm$2.8&94.7$\pm$0.6&97.5$\pm$0.8&97.2$\pm$0.5&73.7$\pm$3.5&88.4$\pm$1.1&98.1$\pm$0.8&97.1$\pm$1.4&87.8$\pm$0.8&97.1$\pm$0.6&95.9$\pm$0.7&91.2$\pm$1.2  \\
			\hline
		  MEDA(s)~\cite{MEDA} &94.4$\pm$1.0&71.2$\pm$3.9&92.8$\pm$0.8&97.3$\pm$0.6&97.4$\pm$0.5&70.6$\pm$3.5&86.3$\pm$1.1&97.4$\pm$1.0&96.2$\pm$1.5&85.6$\pm$1.0&97.6$\pm$0.8&96.5$\pm$0.9&90.3$\pm$1.4  \\
			\hline
			Our(s)            &\textbf{96.0}$\pm$0.5&\textbf{78.5}$\pm$1.8&\textbf{96.8}$\pm$0.5&\textbf{97.6}$\pm$0.4&\textbf{97.6}$\pm$0.4&\textbf{78.5}$\pm$1.2&\textbf{89.0}$\pm$0.6&\textbf{99.1}$\pm$0.6&\textbf{98.9}$\pm$0.5&\textbf{89.1}$\pm$0.6&97.6$\pm$0.6&\textbf{97.4}$\pm$1.0&\textbf{93.0}$\pm$0.7  \\
			\hline
			TCA(t)~\cite{TCA} &76.4$\pm$0.9&74.8$\pm$4.7&91.3$\pm$1.2&87.0$\pm$0.6&73.3$\pm$0.5&82.4$\pm$8.7&73.9$\pm$0.7&96.0$\pm$0.8&95.9$\pm$1.3&76.2$\pm$2.0&64.1$\pm$0.2&63.4$\pm$0.4&79.6$\pm$1.8  \\
			\hline
			JDA(t)~\cite{JDA} &76.2$\pm$0.9&74.1$\pm$6.0&93.7$\pm$0.8&90.1$\pm$0.9&75.3$\pm$1.2&81.8$\pm$7.1&80.4$\pm$1.4&95.4$\pm$1.3&96.2$\pm$1.6&79.1$\pm$2.4&67.2$\pm$0.6&67.5$\pm$0.9&81.4$\pm$2.1  \\
			\hline
			BDA(t)~\cite{BDA} &74.9$\pm$1.0&74.2$\pm$4.1&93.4$\pm$0.7&90.4$\pm$0.9&75.8$\pm$0.9&79.8$\pm$6.3&77.6$\pm$1.4&93.3$\pm$1.4&95.5$\pm$1.3&77.8$\pm$1.4&66.2$\pm$0.9&66.2$\pm$0.8&80.4$\pm$1.8  \\
			\hline
			VDA(t)~\cite{VDA} &76.6$\pm$0.6&78.1$\pm$7.3&93.9$\pm$0.4&91.6$\pm$0.7&75.6$\pm$0.9&85.6$\pm$7.2&80.9$\pm$1.5&96.3$\pm$1.2&96.5$\pm$1.6&79.5$\pm$2.5&68.3$\pm$0.5&68.0$\pm$1.1&82.6$\pm$2.1  \\
			\hline
		  JGSA(t)~\cite{JGSA} &76.1$\pm$0.6&81.2$\pm$5.9&93.6$\pm$0.7&91.4$\pm$0.7&76.3$\pm$0.7&87.9$\pm$6.6&82.8$\pm$1.7&96.5$\pm$0.9&96.8$\pm$1.5&80.9$\pm$1.9&69.7$\pm$0.5&71.2$\pm$0.5&83.7$\pm$1.9  \\
			\hline
		  MEDA(t)~\cite{MEDA} &\textbf{79.1}$\pm$0.5&83.7$\pm$7.4&\textbf{94.8}$\pm$1.6&\textbf{91.9}$\pm$0.6&78.2$\pm$0.8&89.1$\pm$6.8&\textbf{84.6}$\pm$1.4&96.7$\pm$0.7&96.8$\pm$1.5&82.8$\pm$1.0&72.2$\pm$1.0&\textbf{73.6}$\pm$1.3&85.3$\pm$2.1  \\
			\hline
			Our(t)            &78.8$\pm$0.5&\textbf{84.3}$\pm$5.5&\textbf{94.8}$\pm$0.2&\textbf{91.9}$\pm$0.3&\textbf{78.5}$\pm$0.2&\textbf{90.1}$\pm$4.7&82.1$\pm$0.7&\textbf{97.8}$\pm$0.5&\textbf{98.7}$\pm$0.7&\textbf{85.8}$\pm$1.0&\textbf{74.5}$\pm$0.2&71.8$\pm$0.2&\textbf{85.8}$\pm$1.2  \\
			\hline
		\end{tabular}
	\end{center}
\end{table}

\begin{table}[h]
	\begin{center}
		\fontsize{5}{6}\selectfont
		\caption{Accuracy (\%) on the Office-Home dataset with ResNet50 features in SLSA-DA setting} \label{tab6}
		\begin{tabular}{|c|c|c|c|c|c|c|c|c|c|c|c|c|c|}
			\hline
			SLSA-DA
			&Ar$\rightarrow$Cl&Ar$\rightarrow$Pr&Ar$\rightarrow$Rw&Cl$\rightarrow$Ar&Cl$\rightarrow$Pr&Cl$\rightarrow$Rw&Pr$\rightarrow$Ar&Pr$\rightarrow$Cl&Pr$\rightarrow$Rw&Rw$\rightarrow$Ar&Rw$\rightarrow$Cl&Rw$\rightarrow$Pr&Avg.
			\\
			\hline
			\hline
			TCA(s)~\cite{TCA} &71.8$\pm$0.5&72.2$\pm$0.9&72.4$\pm$0.7&73.5$\pm$0.7&73.3$\pm$1.4&73.5$\pm$0.9&87.4$\pm$0.6&87.7$\pm$1.0&87.7$\pm$0.7&82.4$\pm$1.0&81.9$\pm$1.1&82.3$\pm$0.9&78.8$\pm$0.9  \\
			\hline
			JDA(s)~\cite{JDA} &69.3$\pm$0.9&70.6$\pm$1.2&72.5$\pm$0.8&68.8$\pm$1.1&70.2$\pm$1.3&70.9$\pm$0.9&82.7$\pm$0.7&81.6$\pm$1.1&85.2$\pm$0.9&78.9$\pm$1.0&76.4$\pm$1.3&81.2$\pm$0.8&75.7$\pm$1.0  \\
			\hline
			BDA(s)~\cite{BDA} &63.2$\pm$1.0&63.9$\pm$1.4&68.1$\pm$1.1&64.1$\pm$1.0&66.4$\pm$1.6&68.1$\pm$1.2&77.4$\pm$1.0&76.9$\pm$1.2&81.1$\pm$1.1&74.0$\pm$1.2&71.3$\pm$1.3&75.1$\pm$1.0&70.8$\pm$1.2  \\
			\hline
			VDA(s)~\cite{VDA} &71.3$\pm$1.1&72.7$\pm$1.0&74.1$\pm$0.7&72.2$\pm$1.2&72.4$\pm$1.3&73.2$\pm$0.8&84.4$\pm$1.1&83.5$\pm$1.3&86.7$\pm$0.9&80.8$\pm$0.9&78.3$\pm$1.3&82.3$\pm$0.8&77.7$\pm$1.0  \\
			\hline
		  JGSA(s)~\cite{JGSA} &72.7$\pm$0.8&72.8$\pm$1.0&74.0$\pm$0.8&73.8$\pm$0.9&74.4$\pm$1.4&75.0$\pm$1.1&87.9$\pm$0.7&88.4$\pm$0.9&89.0$\pm$0.9&82.9$\pm$0.9&82.3$\pm$1.1&83.9$\pm$0.9&79.8$\pm$1.0  \\
			\hline
		  MEDA(s)~\cite{MEDA} &69.4$\pm$0.9&69.3$\pm$0.7&69.3$\pm$0.6&73.1$\pm$0.7&72.8$\pm$1.1&73.2$\pm$0.8&88.1$\pm$0.6&88.1$\pm$0.8&88.1$\pm$0.8&81.6$\pm$0.9&81.2$\pm$1.0&81.9$\pm$0.7&78.0$\pm$0.8  \\
			\hline
			Our(s)            &\textbf{77.8}$\pm$0.6&\textbf{78.2}$\pm$0.3&\textbf{79.3}$\pm$0.4&\textbf{80.2}$\pm$1.0&\textbf{79.9}$\pm$1.2&\textbf{79.2}$\pm$0.6&\textbf{91.2}$\pm$0.4&\textbf{91.6}$\pm$0.8&\textbf{91.7}$\pm$0.5&\textbf{88.7}$\pm$0.5&\textbf{87.7}$\pm$0.5&\textbf{88.7}$\pm$0.5&\textbf{84.5}$\pm$0.6  \\
			\hline	 
			TCA(t)~\cite{TCA} &42.9$\pm$1.3&57.9$\pm$2.2&62.2$\pm$1.1&46.6$\pm$1.0&55.1$\pm$1.0&57.6$\pm$0.9&49.3$\pm$1.1&42.9$\pm$0.8&68.6$\pm$0.8&59.6$\pm$1.1&48.0$\pm$0.6&71.1$\pm$1.4&55.2$\pm$1.1  \\
			\hline
			JDA(t)~\cite{JDA} &44.7$\pm$1.1&59.4$\pm$1.6&63.7$\pm$1.0&49.0$\pm$1.3&58.4$\pm$1.7&59.1$\pm$1.0&51.1$\pm$1.1&44.9$\pm$1.0&70.0$\pm$0.9&60.3$\pm$1.3&49.4$\pm$1.0&72.4$\pm$1.8&56.9$\pm$1.2  \\
			\hline
			BDA(t)~\cite{BDA} &41.0$\pm$0.8&53.5$\pm$1.6&60.7$\pm$1.2&46.2$\pm$1.5&55.5$\pm$1.7&56.8$\pm$1.3&47.3$\pm$1.6&43.1$\pm$1.0&66.1$\pm$1.3&57.0$\pm$1.1&47.4$\pm$1.0&68.2$\pm$2.0&53.6$\pm$1.3  \\
			\hline
			VDA(t)~\cite{VDA} &45.7$\pm$1.3&61.9$\pm$1.9&65.3$\pm$1.2&50.9$\pm$0.9&60.4$\pm$1.7&60.7$\pm$1.0&53.3$\pm$0.8&46.2$\pm$0.8&71.5$\pm$1.1&62.0$\pm$1.4&50.4$\pm$1.1&74.0$\pm$1.9&58.5$\pm$1.3  \\
			\hline
		  JGSA(t)~\cite{JGSA} &46.2$\pm$1.2&63.4$\pm$2.2&67.3$\pm$1.2&49.8$\pm$1.3&61.5$\pm$2.0&61.5$\pm$1.6&53.3$\pm$0.9&46.6$\pm$0.9&73.1$\pm$0.8&61.1$\pm$1.3&50.3$\pm$1.1&74.1$\pm$1.8&59.0$\pm$1.4  \\
			\hline
		  MEDA(t)~\cite{MEDA} &50.2$\pm$1.1&67.8$\pm$2.2&71.7$\pm$0.9&53.5$\pm$0.9&66.3$\pm$2.0&66.6$\pm$0.9&56.6$\pm$0.7&49.6$\pm$0.9&\textbf{76.0}$\pm$0.8&64.7$\pm$1.7&54.0$\pm$1.2&78.1$\pm$1.0&62.9$\pm$1.2  \\
			\hline
			Our(t)            &\textbf{54.1}$\pm$1.3&\textbf{71.5}$\pm$1.3&\textbf{73.5}$\pm$1.1&\textbf{58.6}$\pm$0.6&\textbf{67.0}$\pm$1.2&\textbf{67.6}$\pm$1.2&\textbf{59.3}$\pm$1.7&\textbf{51.9}$\pm$1.1&75.8$\pm$0.5&\textbf{69.4}$\pm$0.5&\textbf{57.9}$\pm$0.7&\textbf{80.5}$\pm$0.5&\textbf{65.6}$\pm$1.0  \\
			\hline
		\end{tabular}
	\end{center}
\end{table}

\begin{figure}[!h]
	\begin{center}
		\includegraphics[width=1.0\linewidth,height=0.6\textheight]{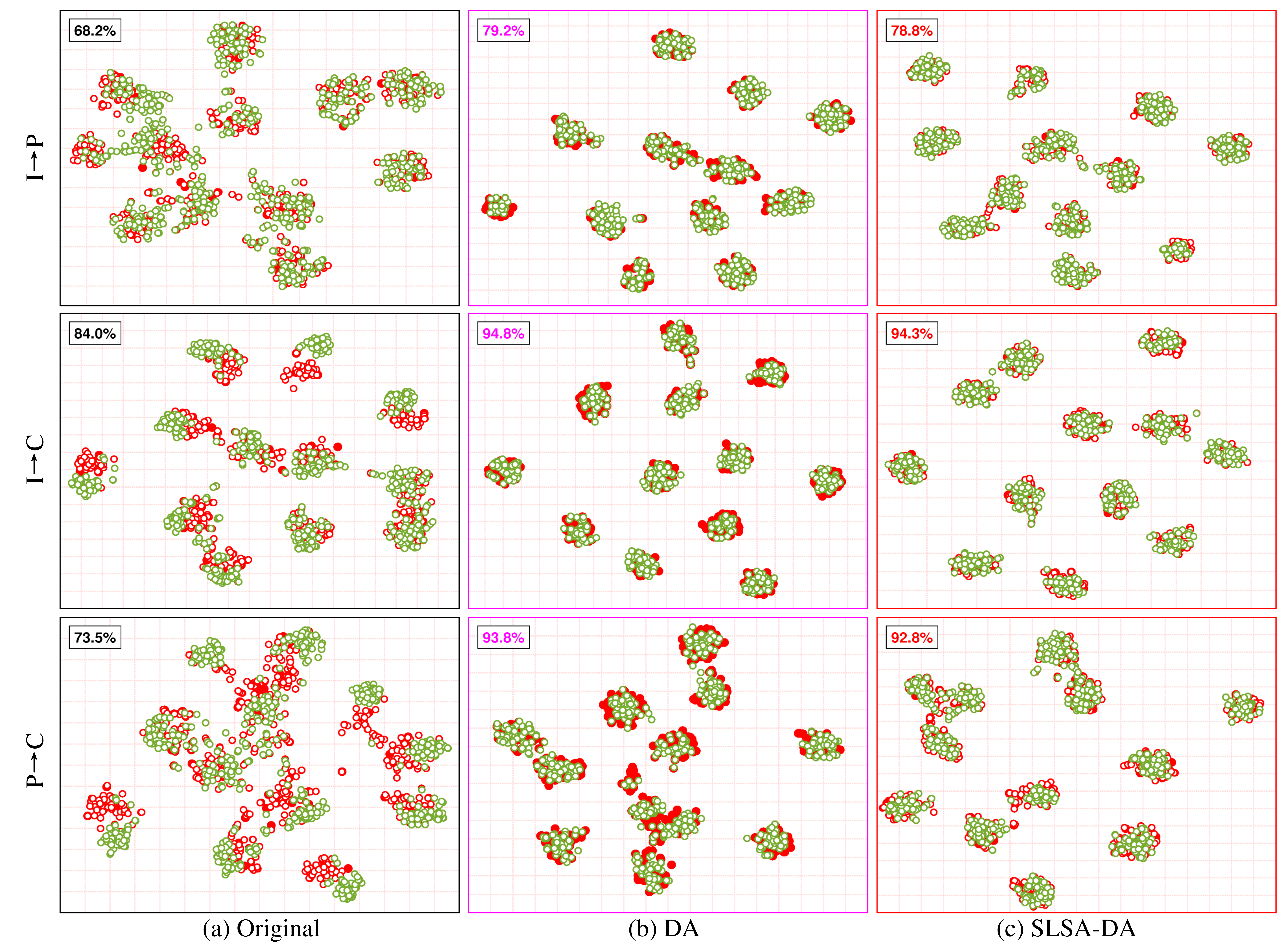}
	\end{center}
	\caption{tSNE visualization on ImageCLEF-DA. The source domain is denoted by red dots and the target domain is denoted by green dots. The figures in black boxes are their corresponding target accuracy. The red solid dots are the labeled source data, while the red and green hollow dots are the unlabeled source and target data, respectively.}
	\label{fig4}
\end{figure}

\begin{figure}[!h]
	\begin{center}
		\includegraphics[width=0.85\linewidth,height=0.4\textheight]{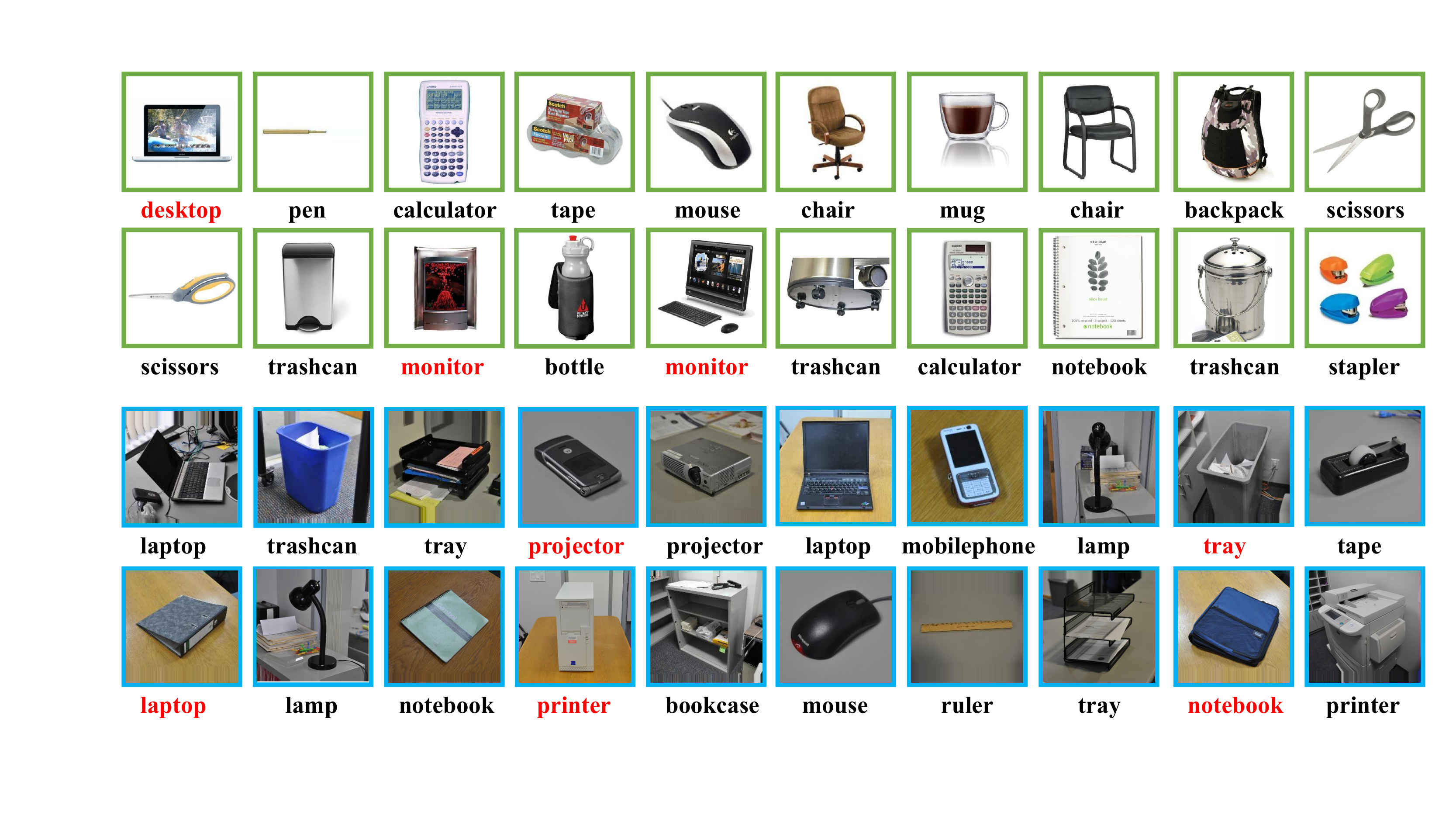}
	\end{center}
	\vspace{-40pt}
	\caption{Recognition results for unlabeled source (green boxes) and target (blue boxes) samples, correct and incorrect labeled instances are marked in black and red fonts, respectively.}
	\label{fig5}
\end{figure}

\begin{figure}[!h]
	\begin{center}
		\includegraphics[width=1.0\linewidth,height=0.45\textheight]{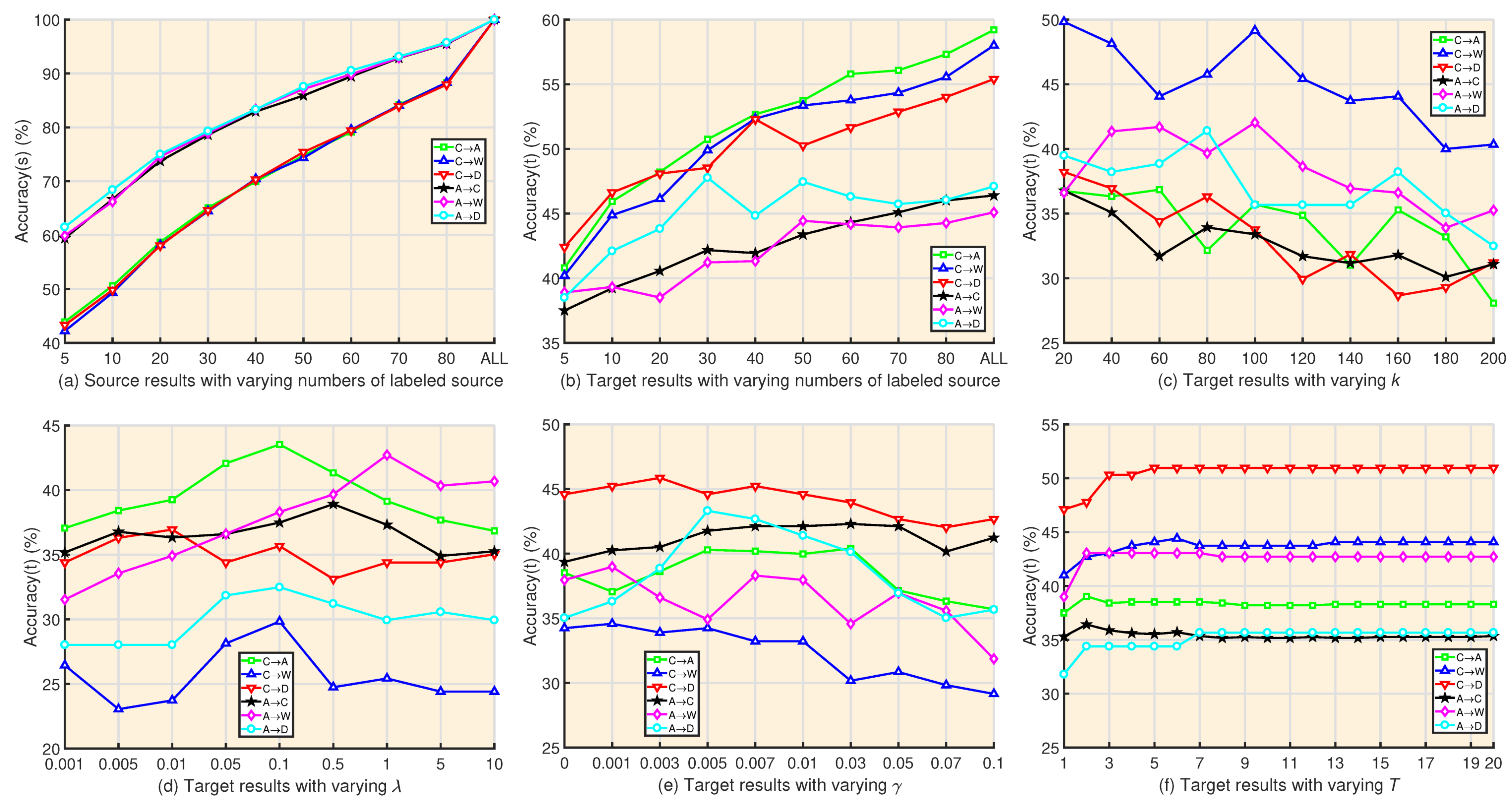}
	\end{center}
	\caption{Effect of the numbers of labeled source instances. Parameter sensitivity, w.r.t. the $k$, $\lambda$, $\gamma$, $T$ values.}
	\label{fig6}
\end{figure}

\par In addition, we employ two evaluation metrics: s: the accuracy for all of the source domain including the labeled data; t: the accuracy for target domain. We implemented those 6 shallow DA methods in SLSA-DA setting, then compared with the proposed approach. For a fair comparison, the parameters of all shallow DA methods are set to same.

\par The performances of different methods in SLSA-DA settings are shown in Table~\ref{tab4}, Table~\ref{tab5}, Table~\ref{tab6}. It can be observed that the accuracy for the target domain are reduced largely, since there are only a few labeled source data. However, our approach outperforms those 6 methods on nearly all of the evaluations, not only the accuracy of source domain, but also target domain, which indicates that our approach could still achieve best results for the SLSA-DA scenario.

\par TCA, JDA, VDA can be viewed as a special case of our approach. From Eq.(\ref{eq8}), TCA is the case with $\textbf{\textit{L}}_{mmd1}$ considered, while JDA take into account both $\textbf{\textit{L}}_{mmd1}$ and $\textbf{\textit{L}}_{mmd2}$. VDA further constructed the intra-class scatter matrix to preserve discriminative information of source domain based on JDA. In addition, BDA introduced a weighted mechanism to deal with class imbalance issues based on JDA. JGSA jointly aligned their geometrical and statistical structures, to tackle with huge distributional discrepancy across the two domains. MEDA realizes distributional alignment in Grassmann manifold to mitigate degenerated feature transformation. However, all of them separate the label prediction and distributional alignment as different steps, and rich source labels are required. Different from them, our approach takes advantage of the projected clustering, label propagation and distributional alignment, and incorporates them into a unified framework seamlessly. Therefore, the proposed approach could achieve best results in both DA and SLSA-DA settings. Specially, the results on Office-Home dataset verify that our method is applicable to the large-scale dataset, where 1.0\% improvement is achieved against MDD in DA setting and 6.6\%(s), 2.6\%(t) over MEDA in SLSA-DA setting.

\subsection{Experimental Analysis}

\par In this subsection, we present experimental analysis of the proposed approach from four aspects. Firstly, we make a visualization of feature representation (Original, DA, SLSA-DA). Secondly, we evaluate the effect of labeled source number from each class. Thirdly, we visualize the recognition results on the unlabeled source and target domains. Finally, we analyze the sensitivity of the proposed approach with respect to hyper-parameters $k$, $\lambda$, $\gamma$, and check its convergence property by $T$.

\subsubsection{Feature Visualization}

\par Following the work in~\cite{JAN, MADA}, we visualize the features learned by the proposed approach with DA and SLSA-DA settings on the tasks I$\rightarrow$P, I$\rightarrow$C, P$\rightarrow$C of the ImageCLEFF dataset. The results of feature visualization for original features, DA features and SLSA-DA features are illustrated in Fig.~\ref{fig4} . Comparing with original features, we observed that not only the performance of inter-domain distributional alignment is improved (i.e., the red dots should be aligned with as many green dots as possible), but also the discriminative ability of each domain is promoted (i.e., the red/green dots should be assembled as tightly as possible, and the cluster numbers equal to the class numbers). Furthermore, the features learned from SLSA-DA are competitive with DA, although there are only a few labeled data instances in the source domain (i.e., the number of solid red dots in SLSA-DA is smaller than DA). Generally speaking, the proposed approach not only achieves better performances in DA, but also in the proposed SLSA-DA setting.

\subsubsection{Labeled Numbers}
\par We conducted experiments on six SLSA-DA tasks from Office10-Caltech10 dataset with different source labeled numbers from each class. As can be seen from Fig.~\ref{fig6}~(a), the recognition accuracy of source domain is increased with the labeled numbers. However, most of the time, the recognition accuracy of target domain is largely or slightly increased, sometimes it is marginally decreased (i.e., Fig.~\ref{fig6}~(b)). We argue that some source data instances might have negative effect on knowledge transfer, although they are labeled correctly. 

\subsubsection{Recognition Results}
\par In SLSA-DA scenario, we conduct the proposed approach on the task A$\rightarrow$D from Office31 dataset, and randomly select 20 source/target images with their predictive results. As shown in Fig.~\ref{fig5}, our method could obtain desirable results in this weaker DA setting, where the correct and incorrect labeled instances are marked in black and red fonts.

\subsubsection{Impact of Hyper-Parameters}

\par The proposed model entails 4 hyper-parameters, and we conduct sensitivity analysis to validate that the optimal results could be achieved under a wide range of parameter values. We only report target recognition results on six SLSA-DA tasks from Office10-Caltech10 dataset, and similar trends on all other evaluations are proved, but not shown here due to space limitation. 

\par We run the proposed approach with varying values of $k$. It can be chosen such that the low-dimensional representation is accurate for data reconstruction. We also run the proposed approach with varying values of $\lambda$. Then, we plot classification accuracy w.r.t., different values of $k$ and $\lambda$ in Fig.~\ref{fig6}~(c) and Fig.~\ref{fig6}~(d), and choose $k\in[20, 200]$, $\lambda\in[0.001, 10]$. Theoretically, larger values of $\lambda$ can make scale control of $\textbf{\textit{A}}$ more important. Therefore, the value of $\lambda$ has to increase with the value of $k$ increased. Moreover, we show classification accuracy w.r.t., different values of $\gamma$ in Fig.~\ref{fig6}~(e), and $\gamma\in[0, 0.1]$, which embodies the importance of projected clustering or discriminative preservation. All in all, the proposed approach displays its stability as the resultant classification accuracy remains roughly the same despite a wide range of $k$, $\lambda$ and $\gamma$ values. We also empirically check the convergence property of the proposed approach. Fig.~\ref{fig6}~(f) shows that classification accuracy increases steadily with more iterations and converges within only a few iterations.

\section{Conclusion}

\par In this paper, we introduce a novel DA scenario, referred to as Sparsely-Labeled Source Assisted Domain Adaptation (SLSA-DA), which is a more realistic situation and still under insufficient exploration so far. With regard to the proposed model, to the best our knowledge, SLSA-DA is the first attempt to incorporate the projecting clustering, label propagation and distributional alignment into a unified optimization framework seamlessly. Specifically, we prove that the class-wise MMD could be rewritten as the cluster-wise MMD when the variables related to the cluster centroids are optimized, while the projected clustering could be reformulated as the intra-class scatter minimization when the shared projection is optimized. Therefore, those three quantities could be optimized in one framework, so that they can take advantage of each other's merits. From the experimental results, we observe that our approach could not only achieve competitive capability in DA setting compared with several state-of-art methods, either the shallow or deep ones, but also reflect its superiority in the respects of SLSA-DA scenarios. However, the proposed approach could not deal with the challenge of label quality, especially when there exist the wrongly-labeled data instances in the source domain. Moreover, SLSA-DA still assumes that the feature and label spaces are shared between the source and target domains, which limits the extension of the method where both feature and label spaces can be varied given the practical setting in domain adaption. As it should be, we will extend the proposed model to deal with more complicate situations in our future work, such as partial wrongly-labeled and sparsely-labeled DA, sparsely-labeled PDA, sparsely-labeled PDA, OSDA and UDA, etc.

\section{Acknowledgements}
This work was supported by National Natural Science Foundation of China (NSFC) under Grant 61976042, 61772108, 61572096, 61733002 and by the Fundamental Research Funds for the Central Universities.

\newpage
\bibliography{mybibfile}

\end{document}